\documentclass{article}

\usepackage{microtype}
\usepackage{graphicx}
\usepackage{subcaption}
\usepackage{booktabs} 
\usepackage{listings}
\usepackage{placeins} 

\usepackage{hyperref}

\usepackage[accepted]{icml2026}

\usepackage{amsmath}

\expandafter\def\expandafter\normalsize\expandafter{%
    \normalsize%
    \setlength\abovedisplayskip{4pt}%
    \setlength\belowdisplayskip{6pt}%
    \setlength\abovedisplayshortskip{6pt}%
    \setlength\belowdisplayshortskip{-5pt}%
}

\usepackage{amssymb}
\usepackage{mathtools}
\usepackage{amsthm}

\usepackage[capitalize,noabbrev]{cleveref}

\theoremstyle{plain}

\theoremstyle{definition}

\theoremstyle{remark}

\usepackage[textsize=tiny]{todonotes}

\icmltitlerunning{Reasoning Theater: Disentangling Model Beliefs from Chain-of-Thought}

\begin{document}

\twocolumn[
  \icmltitle{Reasoning Theater: Disentangling Model Beliefs from Chain-of-Thought}


  \icmlsetsymbol{equal}{*}

  \begin{icmlauthorlist}
    \icmlauthor{Siddharth Boppana}{equal,comp}
    \icmlauthor{Annabel Ma}{equal,sch}
    \icmlauthor{Max Loeffler}{comp}
    \icmlauthor{Raphael Sarfati}{comp}
    \icmlauthor{Eric Bigelow}{comp}
    \icmlauthor{Atticus Geiger}{comp}
    \icmlauthor{Owen Lewis}{comp}
    \icmlauthor{Jack Merullo}{comp}
  \end{icmlauthorlist}

  \icmlaffiliation{comp}{Goodfire AI}
  \icmlaffiliation{sch}{Harvard University, Cambridge, MA}

  \icmlcorrespondingauthor{Siddharth Boppana}{sid.boppana@goodfire.ai}
  \icmlcorrespondingauthor{Annabel Ma}{annabelma@college.harvard.edu}

  \icmlkeywords{Machine Learning, ICML}

  \vskip 0.3in
]



 \printAffiliationsAndNotice{\icmlEqualContribution}

\begin{abstract}
We provide evidence of \textit{performative} chain-of-thought (CoT) in reasoning models, where a model becomes strongly confident in its final answer, but continues generating tokens without revealing its internal belief. 
Our analysis compares activation probing, early forced answering, and a CoT monitor across two large models (DeepSeek-R1 671B \& GPT-OSS 120B) and find task difficulty-specific differences: The model's final answer is decodable from activations far earlier in CoT than a monitor is able to say, especially for easy recall-based MMLU questions. We contrast this with genuine reasoning in difficult multihop GPQA-Diamond questions. Despite this, \textit{inflection points} (e.g., backtracking, `aha' moments) occur almost exclusively in responses where probes show large belief shifts, suggesting these behaviors track genuine uncertainty rather than learned ``reasoning theater." Finally, probe-guided early exit reduces tokens by up to 80\% on MMLU and 30\% on GPQA-Diamond with similar accuracy, positioning attention probing as an efficient tool for detecting performative reasoning and enabling adaptive computation.\footnote{Codebase: \href{https://github.com/AskSid/disentangling-computation-from-cot}{github.com/AskSid/disentangling-computation-from-cot}}\footnote{Visualization of our data: \href{https://reasoning-theater.streamlit.app}{reasoning-theater.streamlit.app}}

\end{abstract}

\section{Introduction}

\begin{figure*}[t]
    \centering
    \includegraphics[width=0.92\textwidth]{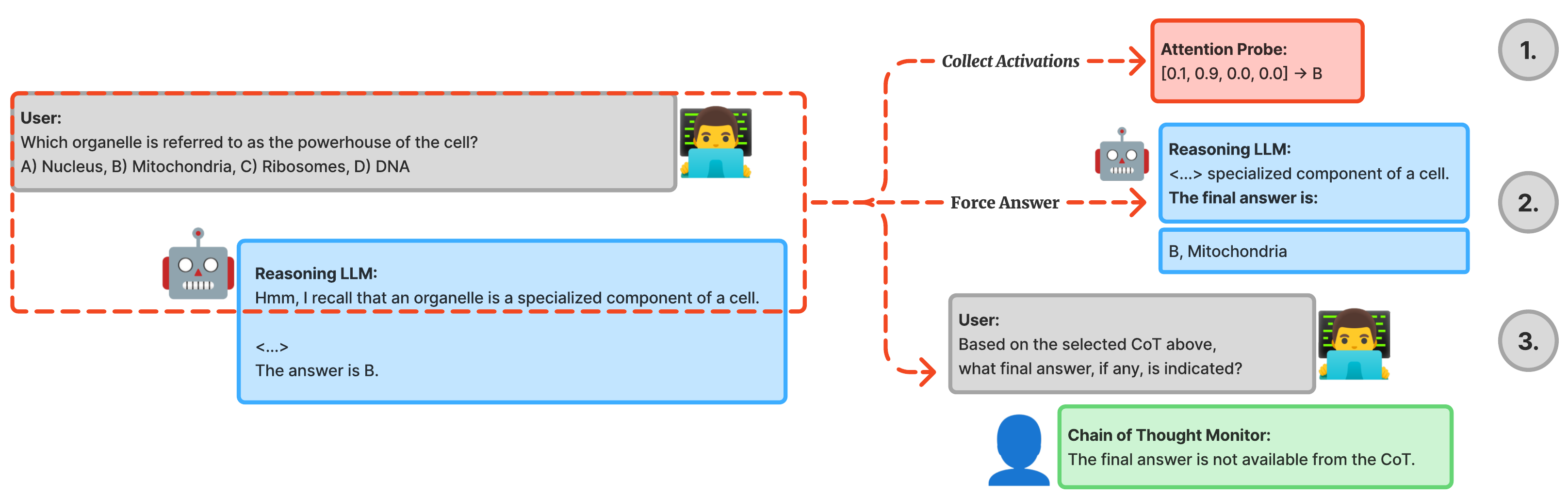}
    \caption{\textbf{Early decoding helps us identify performative reasoning, when an LLM knows what it will answer.} We study whether a reasoning LLM's final answer can be decoded given a prefix of its chain of thought up to an intermediate token $x$. We use this to identify \emph{performative reasoning}, where a model internally knows its final answer early on but still generates text as if it does not. (1) Attention Probes: We train attention probes on varying-length activations of text to predict the model's final answer. At test-time, we use activations up to $x$ to study \textit{when} the model internalizes its final answer. (2) Forced Answering: At token $x$, we inject a forced answering prompt to obtain its final answer prediction at that point in reasoning. (3) Chain-of-thought Monitor: We provide the chain of thought up to $x$ to another LLM, which determines whether the reasoning chain contains a potential final answer.}
    \label{fig:methods-overview}
\end{figure*}

Chain-of-thought (CoT) has emerged as a powerful method to improve Large Language Model (LLM) performance by leveraging step-by-step computation, particularly on math and science tasks that require complex reasoning~\cite{reynolds2021prompt, wei2022chain}. Notably, \emph{reasoning} LLMs trained using reinforcement learning (RL) have demonstrated impressive gains, accounting for many of the latest breakthroughs in capabilities~\cite{guo2025deepseekr1arxiv, openai2024learning}. On one hand, long CoTs, which emerge from RL training, appear to provide a promising opportunity for safety and interpretability: if we can watch the model think, we should be able to find evidence of malicious intentions or flawed logic and respond accordingly~\cite{korbak2025chain}. On the other hand, recent research has shown that CoT traces are not necessarily faithful to the internal reasoning processes employed by the model~\cite{turpin2023language, lanham2023measuring, arcuschin2025cotwild, chen2025reasoningmodelsdontsay}, undermining the reliability of these explanations for safety applications. In this work, we study unfaithfulness through the lens of \emph{performative} chain-of-thought~\citep{palod2025performative}, which is a mismatch between external and internal deliberation: the model continues to generate tokens that read as step-by-step reasoning without disclosing its internally committed confidence.

Our results show that models are at times strongly confident in their final answer immediately into reasoning, and this belief can be probed from the activations of the model far before any confidence is verbalized in CoT. We describe this as performative reasoning, as it is unfaithful to the model's confident underlying belief. The full story is complicated, however. Harder tasks that require test-time compute exhibit genuine reasoning, for which this mismatch is not present. We also find that \textit{inflection points} like backtracking, sudden realizations (`aha' moments), and reconsiderations appear almost exclusively in questions that also contain large shifts in internal confidence within CoT reasoning, signifying \textbf{faithful} expressions of internal uncertainty resolution. Our probes are able to distinguish both of these phenomena from performative CoT; this degree of calibration has the added benefit of giving us an early exit signal that generalizes to a task the probes weren't trained on (\S\ref{sec:early-exit}).

We argue CoT monitors are at best cooperative listeners, but reasoning models are not cooperative \textit{speakers} \citep{grice1975logic}, and many failures in CoT faithfulness can be explained by this framing. Thus, we suggest caution in assuming the settings CoT monitors are effective, and propose an activation monitoring strategy that helps cover the failure cases we describe. We make the following contributions:

\begin{enumerate}
    \item \textbf{We propose activation attention probes as an effective way to decode concepts from long-form chain-of-thought reasoning.} 
    We train these probes to predict the model's final answer and evaluate them throughout generation to track how the model's belief state evolves over time (\S\ref{sec:probes}).  
    \item \textbf{Evidence for difficulty-dependent performative reasoning.}
    We find a difficulty-dependent split, extending prior work \citep{palod2025performative}. On MMLU-Redux, CoTs are often performative: answer information is available well before a CoT monitor indicates a conclusion, as opposed to on GPQA-Diamond, which exhibits more genuine reasoning. As model size increases, a similar trend arises: smaller, less capable models require more test-time compute to reach a final answer (\S\ref{sec:results_task_difficulty}).

    \item \textbf{Inflection points correspond with internal uncertainty at the question level.}
    We find that not all extended reasoning is performative: inflection points corresponding to backtracking or `aha' moments are consistent indicators of genuine belief updates. Responses that have consistent high internal confidence contain fewer inflections, but we find inconsistent results on whether inflections coincide with belief updates (\S\ref{sec:results_inflections}).
    \item \textbf{Practical token savings via calibrated early exit.}
    We demonstrate that our attention probes trained on reasoning prefixes are well-calibrated, enabling confidence-based early exit that substantially reduces generation without sacrificing accuracy. Early exit saves 80\% and 30\% of generated tokens on MMLU-Redux and GPQA-Diamond, respectively, with comparable performance (\S\ref{sec:early-exit}).
\end{enumerate}

\section{Related Work}

\textbf{Faithfulness in language model chain-of-thought explanations.} A growing body of work has investigated the faithfulness of chain-of-thought, showing that models can omit the true causes of their decisions. Models can produce plausible but unfaithful rationales, both under explicit interventions and natural settings~\cite{turpin2023language, lanham2023measuring, arcuschin2025cotwild, agarwal2024faithfulness}. This has direct implications for safety proposals that rely on monitoring CoT, which has been framed as a promising but fragile opportunity for detecting misalignment~\cite{korbak2025chain}. Subsequent work has stress-tested CoT monitoring and identified settings where solely relying on textual reasoning fails to reveal critical internal information~\cite{arnav2025cotredhanded, baker2025monitoring, wang2025thinking, emmons2025chain}, identified brittle correlation between CoT length and problem difficulty~\citep{palod2025performative}, as well as explored whether misaligned or high-risk behavior can be predicted before a model finishes reasoning~\cite{chan2025predictalignmentbeforefinishthinking}. Other work has shown that monitoring may be well-suited for faithfulness~\cite{kadavath2022languagemodelsmostlyknow, mayne2026positivecasefaithfulnessllm}, finding that explanations have predictive power.

Although unfaithful CoT is well-documented, predicting how and when it will be unfaithful continues to be difficult. Traditional interpretability tools are ill-equipped to interpret information flow along the sequence dimension, focusing instead on single token activations or layer-wise signals which scale poorly to very long contexts~\cite{wang2022interpretability, dunefsky2024transcoders, ameisen2025circuit}. Black-box approaches study longer responses directly via the output text, such as through CoT monitors~\cite{emmons2025chain, arnav2025cotredhanded} or resampling~\cite{bigelow2025forkingpaths, bogdan2025thoughtanchors, macar2025thought}. While resampling can be an effective causal analysis tool, thorough analysis of even simple behaviors can be cost-prohibitive, especially for models in the hundreds of billions of parameters, or with very long CoTs, like those studied here.

\textbf{Activation Probing.} Activation probing is a practical way to extract latent information from a model's internals by training lightweight linear classifiers on residual stream activations~\cite{alain2016understanding,belinkov2022probing}. Recent work uses probes for safety applications~\cite{mckenzie2025detectinghighstakesinteractionsactivation}, predicting future model behavior and self-verification~\cite{zhang2025knowwhenright, afzal2025knowing, ashok2026language}, and analyzing hidden uncertainty and branching dynamics during generation~\cite{zur2025road, pmlr-v235-ahdritz24a}. Prior work has examined when sparse or structured probes meaningfully improve interpretability and monitoring~\cite{pmlr-v267-kantamneni25a}. Our use of attention pooling probes, which uses attention to pool a set of activations together, builds on this literature by tracking answer information over the course of a reasoning trace and evaluating probe calibration for early exit.

\textbf{Reasoning Model Interpretability.} Recent work in reasoning model interpretability has sought to understand the internal mechanisms underlying chain-of-thought computation. Sentence-level analysis has identified which reasoning steps are causally important for final answers~\cite{bogdan2025thoughtanchors}, while comparative studies have evaluated differences between base and reasoning models, suggesting that reasoning models repurpose pre-trained knowledge~\cite{venhoff2025basemodelsknow}. Our work complements these approaches by examining information flow at the token level throughout generation, tracking when answer representations emerge relative to their expression in the reasoning trace.

\textbf{Cooperative Communication}
\citet{grice1975logic} propose \textit{maxims} of communication that effective and pragmatic communicators abide by. It is interesting to consider CoT monitoring from this perspective. Prior work models informative interactions between speakers and listeners  under rational speech acts (RSA) \citep{frank2012predicting, goodman2013knowledge} and connects this to linguistic pragmatics \citep{goodman2016pragmatic, levinson1983pragmatics}. Currently, reasoning models are perhaps only indirectly optimized for effective/pragmatic communication (if at all), and it may be productive to understand monitoring shortcomings from this perspective, as well as a way to improve monitorability. We touch on these principles here, but future work may develop this framing further.

\begin{figure*}[ht!]
    \centering
    \includegraphics[width=0.75\textwidth]{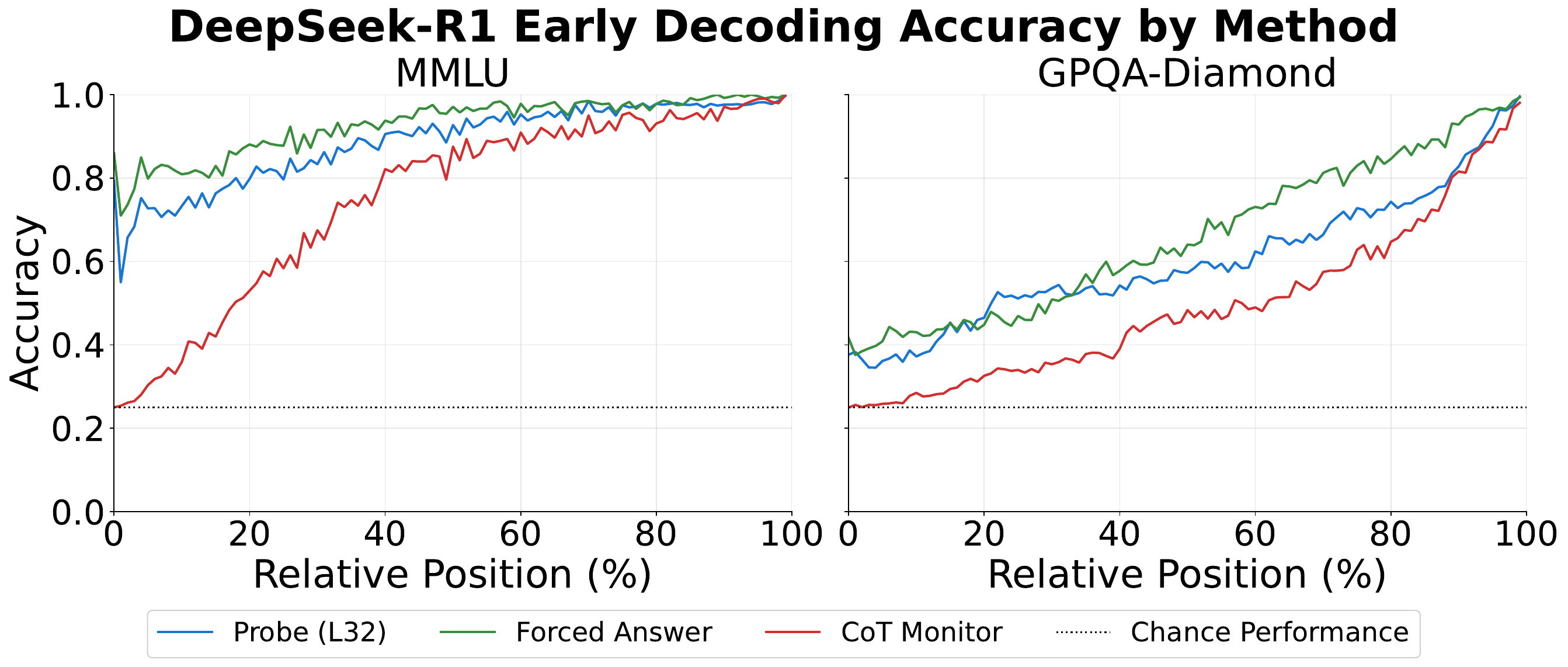}\\[0.5em]
    \includegraphics[width=0.75\textwidth]{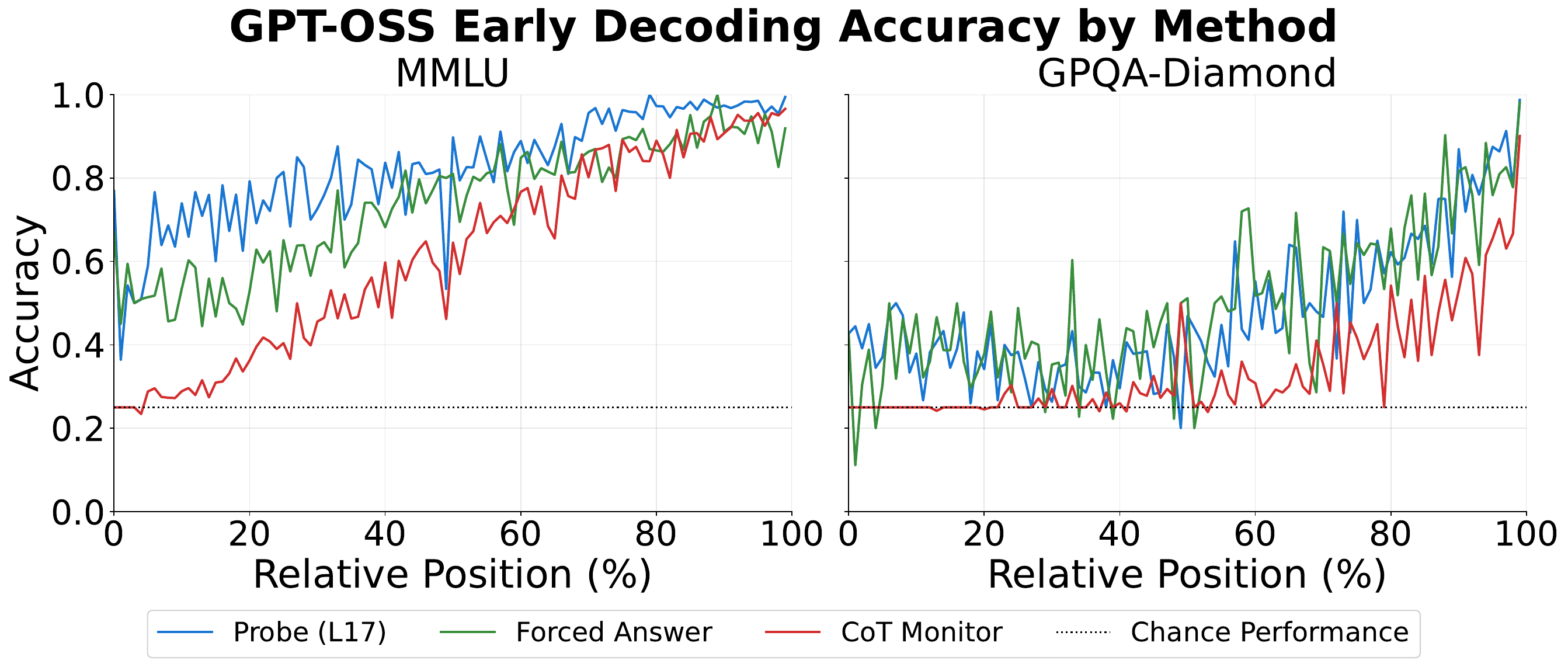}
    \caption{\textbf{Accuracy of three early decoding methods by position of DeepSeek-R1 and GPT-OSS on MMLU-Redux and GPQA-Diamond.} \textit{MMLU (left):} For both models, probing and forced answering predict the models' predictions with much higher accuracy earlier than CoT Monitoring. The CoT monitor's accuracy rapidly gains relative to the other two methods, indicating performative CoT that did not lead to the model's internal accuracy improving. \textit{GPQA-D (right):} All three methods begin with similar accuracy around chance performance, and generally increase at similar rates, indicating closer tracking between internal beliefs and CoT. This is genuine, as the CoT generated corresponds to gains in performance seen in probes and forced answering.}
    \label{fig:accuracy-three-methods-all}
\end{figure*}

\section{Methods}

We use three methods to determine when models know their final answer during a reasoning trace, each with different levels of privileged information from generation: probes trained on layer activations, forced answer prompting, and a CoT monitor. Each method predicts the model's final answer from some prefix of reasoning, making it possible to track when answer information can be decoded exclusively from linear directions in activation space, a few extra forward passes, or the response text itself. A high level overview of the three methods is shown in Figure \ref{fig:methods-overview}.

\subsection{Models \& Datasets}

We focus our CoT analysis primarily on DeepSeek-R1-0528 671B~\cite{guo2025deepseekr1arxiv} and GPT-OSS 120B~\cite{agarwal2025gpt}, two open-weight reasoning models with frontier performance on verifiable domains such as math and science. We additionally include analysis on the set of distilled DeepSeek-R1 Qwen2.5 models (1.5B, 7B, 14B, 32B) to study the effect of model size and capability on performative reasoning.

We evaluate these reasoning models on MMLU-Redux 2.0~\cite{gema2025mmlu} and GPQA-Diamond~\cite{rein2023gpqa}, two popular benchmarks requiring both domain-specific knowledge and step-by-step reasoning skills. MMLU-Redux-2.0 is composed of 5700 questions across 57 domains, which we filter down to 5280 questions using provided error annotations. GPQA-Diamond is a harder subset of the GPQA dataset, containing 198 questions requiring graduate-level expertise in biology, chemistry, and physics. These benchmarks are both in multiple-choice format with four options (A-D), allowing us train probes with a classification objective across answer choices rather than decoding the exact semantic answer.

We collect model responses on both datasets using the same inference settings reported by the original authors, and include these details in Appendix~\ref{appendix-inference-details}. When doing analysis at the step level, the reasoning portion of the response (between the $<$think$>$ and $<$/think$>$ tags) is split into paragraphs using the `\textbackslash n\textbackslash n' delimiter.
To train probes, we collected activations from every layer at every token of the responses.

\subsection{Attention Probes}

We use the attention probe introduced by \citet{pmlr-v267-kantamneni25a}, which applies a learned pooling over a transformer layer's hidden states. During training, we sample random prefixes of the full sequence to encourage the probe to decode the final answer independent of sequence length. At inference time, we evaluate the probe at every prefix position, yielding a probability distribution over answer choices that evolves throughout generation. For each model, we train one probe per layer; results are reported using the best-performing layer across all sequence positions (all layers reported in Appendix \ref{appendix-heatmaps}).

\subsection{Forced Answering}

Forced answer prompting (i.e., early answering) is a method that has been used for validating whether models rely on their chain-of-thought reasoning~\cite{lanham2023measuring, wang2026thinkingcheatingdetectingimplicit}. Given a reasoning trace, we truncate at some intermediate step and prompt the same model to provide its final answer choice (A-D), bypassing the remaining steps. We gather predictions by taking the logits for the four classes and computing the softmax over them to obtain a probability distribution. The model's forced answer has access to all layers' activations from previous text, and can aggregate information to make its final prediction as there are a few forward passes before the final answer is generated. We include our forced answering prompt in Appendix~\ref{appendix-forced-answer-prompt}.

\subsection{Chain of Thought Monitoring}
\label{sec:cot_monitoring_methods}

A CoT monitor is a LLM for evaluating another model's response, often used in safety settings to detect misaligned behavior~\cite{korbak2025chain}. We use a CoT monitor for two distinct purposes:

\paragraph{Predicting the final answer:} Given a question, our CoT monitor is prompted to predict whether the reasoning model has committed to a final answer from a prefix of reasoning steps. The monitor either predicts one of the four choices, or outputs `N/A' if the partial CoT is not sufficient to predict the final answer. Allowing the monitor to defer to `N/A' de-incentivizes it from computing the answer based on its own knowledge.

\paragraph{Identifying inflection points:} We additionally (separately) prompt the CoT monitor to identify key inflection steps in responses, looking for three types: backtracking, realizations, and reconsiderations, and record the steps at which they occur.

We use Gemini-2.5-Flash as our CoT monitor~\cite{comanici2025gemini} and provide both prompts in Appendix~\ref{appendix-cot-monitors-details}.


\section{Attention Probe Results}
\label{sec:probes}

We train attention probes per layer of both DeepSeek-R1 and GPT-OSS, along with the DeepSeek-R1 Qwen distilled models on MMLU questions, and evaluate on a held-out set of MMLU (N=528) and GPQA-D (N=157). We experiment with both direct transfer and fine-tuning on a set of 20 questions, and find negligible improvement for fine-tuning in terms of overall probe accuracy.

Traditional linear probes fail at this task, performing near chance across layers and positions (Appendix~\ref{appendix-baseline-comparison}). A single token's activation is unlikely to consistently encode the final answer across arbitrary positions in a long reasoning trace, as the belief state updates dynamically at specific tokens. On the other hand, attention probes pool across the sequence dimension, allowing relevant token representations to be weighted higher, and succeed where linear probes do not. We find a distinct relation between layer and final answer prediction accuracy: the second half of layers in DeepSeek-R1 and the last three quarters of GPT-OSS layers can decode the final answer (Appendix~\ref{appendix-heatmaps}).

\section{Easier Tasks Exhibit More Performative Reasoning}
\label{sec:results_task_difficulty}
Probes and forced answering provide us with final answer predictions based on the model's internals, while also providing a level of confidence in their predictions. A CoT monitor measures whether the model has indicated a final answer in its CoT up to a certain step. Comparing these two types of methods allows us to study performative CoT in two ways. If probes and forced answering are much more accurate than the monitor at a specific timestep, the model has not revealed its current belief. Informally, we define performativity as this gap. We find that harder tasks exhibit \textit{genuine} reasoning, where the CoT is necessary and probe confidence and CoT monitor accuracy increase together. Meanwhile easier tasks are much more performative, where a spike in probe or forced answer confidence early in the CoT precedes the monitor reaching its predictions.

\subsection{Dataset Dependent Trends}

\begin{table}[tb]
  \caption{\textbf{Comparison of the amount of information gained per step for the CoT Monitor vs. the Probe/Forced Answer.} A large difference indicates that the LLM does not produce its answer based on the addition of information from the CoT while small differences indicate more genuine reasoning. This is measured as the change in slope of the average probe (or forced answer) accuracy minus CoT accuracy shown in Fig. \ref{fig:accuracy-three-methods-all}: $|\Delta \text{Probe}  - \Delta \text{Monitor}|$}
  \label{tab:max_area_between_curves}
  \begin{center}
    \begin{small}
      \begin{sc}
        \begin{tabular}{lcc}
          \toprule
          Model / Dataset
          & \begin{tabular}[c]{@{}c@{}} Probe vs. \\  Monitor\end{tabular}
          & \begin{tabular}[c]{@{}c@{}}Forced vs. \\ Monitor\end{tabular} \\
\midrule
  DeepSeek-R1 (MMLU)       & 0.417 & 0.505 \\                                         
  DeepSeek-R1 (GPQA-D)       & 0.012 & 0.010 \\    
  \midrule
  GPT-OSS (MMLU)           & 0.435 & 0.334 \\                                         
  GPT-OSS (GPQA-D)           & 0.227 & 0.185 \\
          \bottomrule
        \end{tabular}
      \end{sc}
    \end{small}
  \end{center}
  \vskip -0.1in
\end{table}

Figure~\ref{fig:accuracy-three-methods-all} shows the relative accuracies of our three early decoding methods across both DeepSeek-R1 and GPT-OSS, and MMLU and GPQA-D. We collect predictions by step between the $<$think$>$ and $<$/think$>$ tokens in a response.

We find that much of the initial reasoning done in chain-of-thought is performative across both models, with both probes and forced answering showing high accuracy from the beginning of reasoning, while the CoT monitor cannot identify the indicated answer until later. This gap is significantly larger for MMLU than GPQA-D, likely because MMLU questions mainly require recall rather than multi-hop reasoning. For GPQA-D, accuracy for all three methods increases gradually over the reasoning trace, with probes and forced answering still ahead but starting much lower. This shows the necessity of CoT for answering harder questions, in which the rate of information gain per reasoning step is reflected by the probe and CoT monitor, while easier questions exhibit much more performative reasoning.


\paragraph{Performativity Rate} To quantify how much each reasoning step increases probe or forced answer accuracy relative to the CoT monitor, we compute the difference in slopes between the two types of methods at each 5\% timestep bin, averaged over questions and bins. We apply a quadratic fit before computing slopes to smooth over response-count variance, which is particularly pronounced for GPT-OSS due to having fewer reasoning steps overall.

Table~\ref{tab:max_area_between_curves} shows average performativity rate across models and datasets. Under this measurement, both the probe and forced answer show that MMLU is much more performative than GPQA-D: in R1, MMLU has a performativity measure of 0.417, whereas GPQA-D is 0.012. A rate near 0 means added tokens roughly translate to similar increases in both probe and monitor accuracy, indicating genuine reasoning. This trend holds for both models and for forced answering, confirming that MMLU performance does not improve with more tokens relative to the monitor, while GPQA-D shows matched incremental gains. Together, the initial difference in final answer decoding accuracy along with the difference in accuracy slope over sequence length between probes/forced answering and the CoT monitor shows a clear trend of performative CoT for MMLU: models know their final answer early, and do not benefit from additional CoT compared to a monitor.

\subsection{Model Size Dependent Trends}

If harder tasks lead to more faithful CoT, are smaller models more faithful than large models on equivalent datasets? We conduct the same analysis on the DeepSeek-R1 model family, comparing our three early decoding methods on each distilled model's MMLU responses. We exclude GPQA-D from this analysis as the task is too difficult for smaller models, causing answer choice collapse that confounds early decoding.

\begin{figure}[ht]                                                                             
      \centering                         
      \includegraphics[width=\columnwidth]{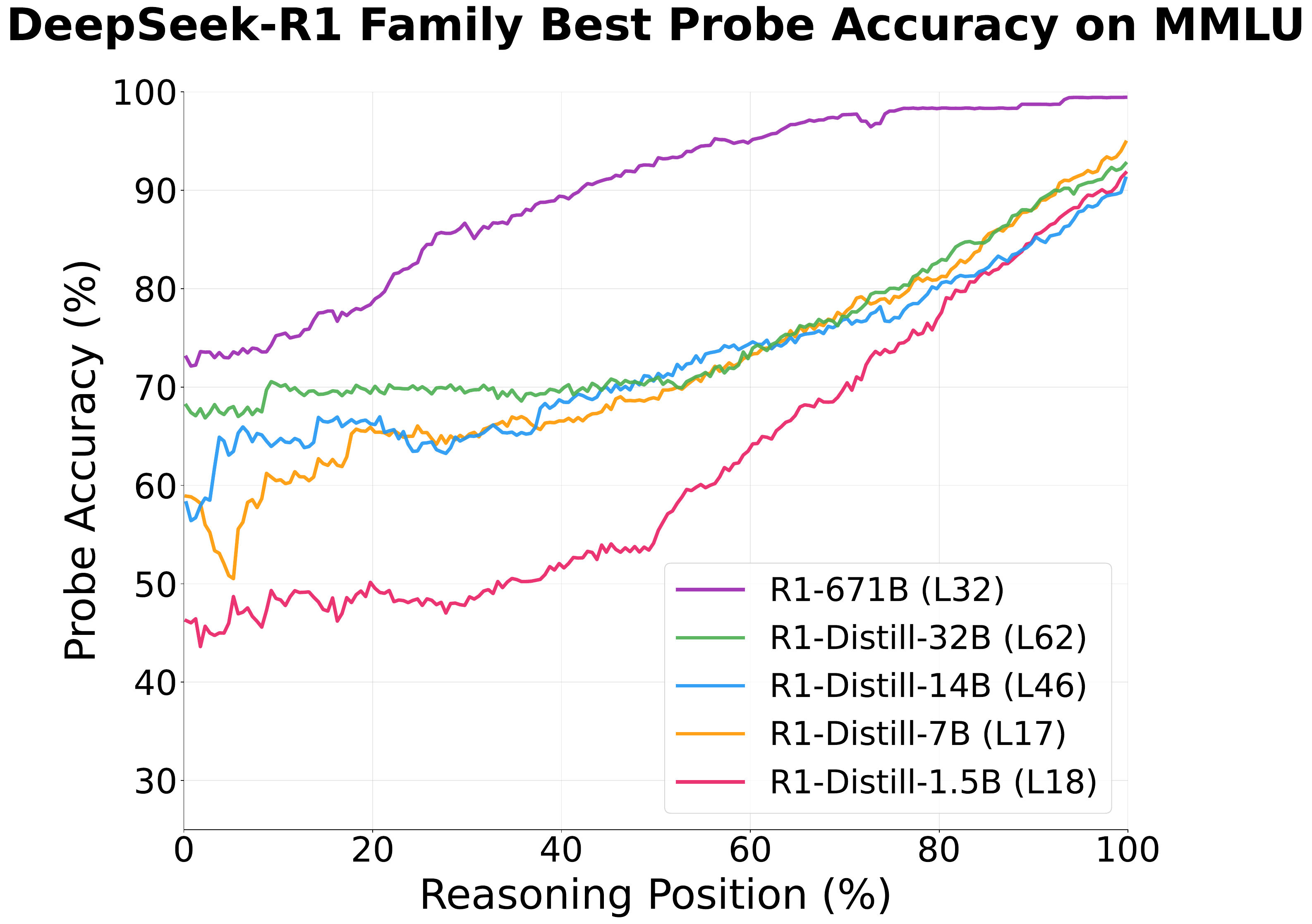}            
      \caption{\textbf{Probe accuracy across DeepSeek-R1 model sizes on MMLU-Redux.} Larger models achieve higher probe accuracy earlier in reasoning, likely
  reflecting greater in-weights knowledge. The 671B probe reaches high accuracy quickly and plateaus, the distilled models (7B--32B) remain flat before rising  
  late in the sequence, and the 1.5B probe starts near chance and sharply increases only in the second half of reasoning.}
      \label{fig:distill_probe_comparison}
  \end{figure}

Figure \ref{fig:distill_probe_comparison} shows probe accuracy over reasoning for each distill along with the original R1 model used earlier. We find that in MMLU responses the final answer can be decoded with greater accuracy at the beginning of reasoning as model size increases. All models converge towards perfect accuracy during reasoning, with the 1.5B model's best probe reaching it at a slower rate in the first half of reasoning. Surprisingly, performance for the 7B, 14B, and 32B model's best probes are largely similar, suggesting comparable ability to decode the final answer information in this model size regime. 

\begin{figure*}
    \centering
    \includegraphics[width=\textwidth]{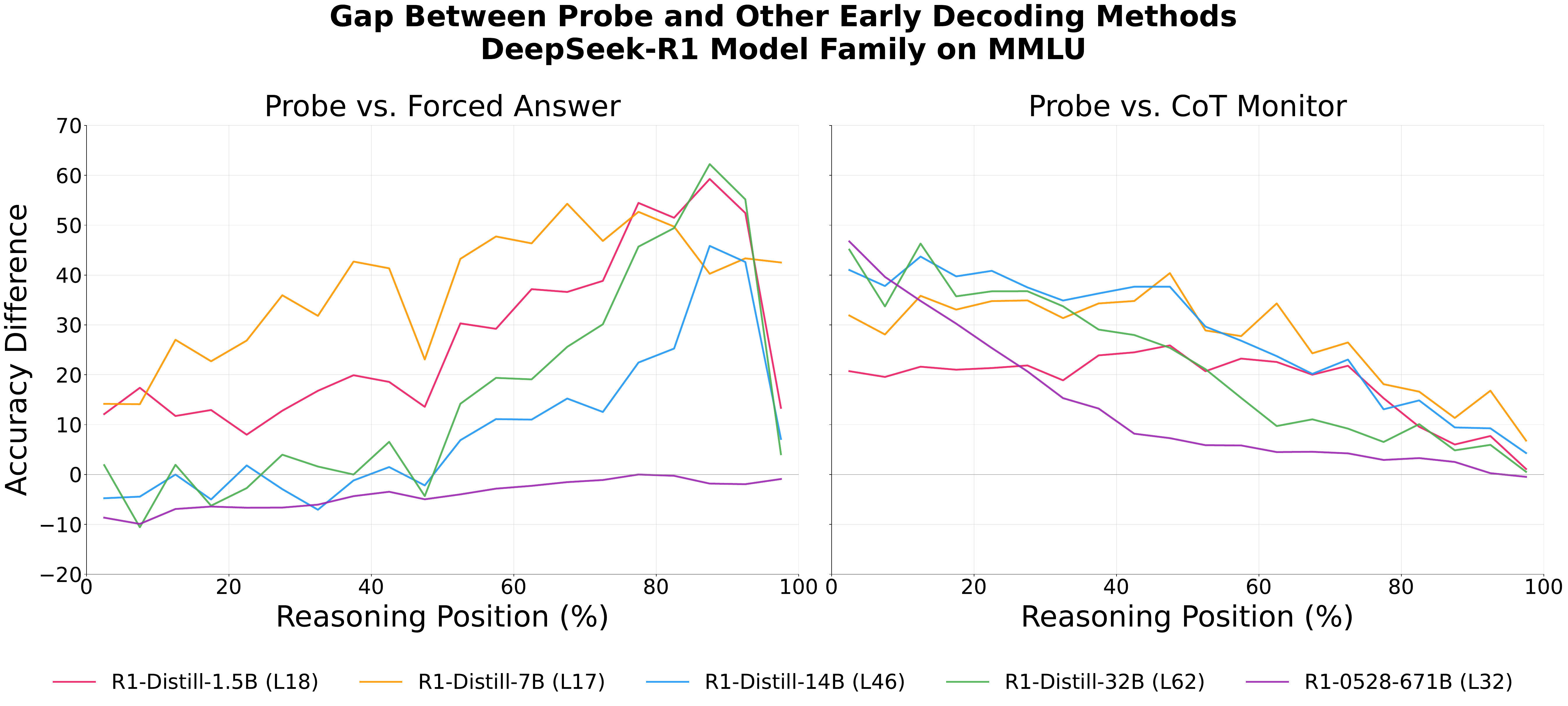}
    \caption{\textbf{Probe performance vs. other methods by model size.} When comparing probes to forced answering, we see that the gap between them rises during reasoning for distilled models due to probe performance increasing while forced answering performance stays the same. In contrast, the gap between probes and the CoT monitor decreases over reasoning. We find that smaller models are less performative as the task is more difficult, requiring genuine reasoning.}
    \label{fig:probe_vs_method_gap}
\end{figure*}

Comparing to forced answering and CoT monitoring (Figure \ref{fig:probe_vs_method_gap}), forced answering accuracy improves relative to probe accuracy as model size increases, potentially due to the off-policy nature of the prompting scheme. All model sizes are able to eventually decode the final answer towards the end of the reasoning trace. Smaller models show a smaller probe-to-monitor gap, particularly in the first half of reasoning, suggesting they generate more faithful CoT relative to their internal beliefs due to a weaker prior on answer choices. However, we note that the 671B DeepSeek-R1's probe performance gap with the CoT monitor rapidly drops relative to smaller models, implying that the CoT ``catches up" to internal beliefs faster over reasoning. Model size seems to be correlated with performative reasoning, with smaller models needing to employ more test-time compute to solve the same problem. We include comparisons of the three decoding methods' accuracy per distilled model separately in Appendix~\ref{appendix-distilled-model-three-methods}.  

\section{Inflection Points Suggest Faithful Reasoning}
\label{sec:results_inflections}
We have shown evidence for performative CoT in the sense that models do not convey that they are highly confident. Here, we investigate the extent to which individual steps betray this. We look at \textit{inflection points} in reasoning like backtracking \citep{guo2025deepseekr1arxiv, ward2025reasoning, venhoff2025understanding} or `Aha' realizations as points of interest. If the model produces these even when it is highly confident internally, these would be highly unfaithful. We have the same CoT monitor label each reasoning step as one of: backtracking, realizations (`aha's), reconsiderations, or not an inflection (see \S \ref{sec:cot_monitoring_methods}) and analyze their occurrences compared to probe confidences.

  \begin{table}[t]
    \caption{\textbf{Inflection points by probe confidence level for DeepSeek-R1 on MMLU (probe layer 60).} Highly confident CoTs produce dramatically fewer inflection points per step, suggesting that when they
  \textit{do} show up, they are genuine.}
    \label{tab:reasoning-events-confidence-r1}
    \begin{center}
      \begin{footnotesize}
        \begin{sc}
          \begin{tabular}{lccc}
            \toprule
            Event Type
            & \begin{tabular}[c]{@{}c@{}} High \\ Conf. \end{tabular}
            & \begin{tabular}[c]{@{}c@{}} Non-High \\ Conf. \end{tabular} \\
            \midrule
            Reconsideration & 0.015 & 0.033 \\
            Realization      & 0.004  & 0.009  \\
            Backtrack     & 0.001 & 0.003  \\
            \midrule
            \textbf{Total} & \textbf{0.020} & \textbf{0.045} \\
            \bottomrule
          \end{tabular}
        \end{sc}
    \end{footnotesize}
    \end{center}
    \vskip -0.1in
  \end{table}

\subsection{Inflection Points Occur in Less Confident Responses}
\label{sec:qualitative}
\paragraph{Setup} First, we label reasoning steps across traces as \textbf{High Confidence} if the probe begins the reasoning trace at 90\% or above confidence in its final answer, and never falls below that threshold throughout the CoT. This criterion basically selects for the most performative CoTs. For this experiment, we look at R1's MMLU traces, which are the most performative, and find 215/522 are High Confidence. 

\paragraph{Results} Table~\ref{tab:reasoning-events-confidence-r1} shows the distribution of inflections across confidence levels, controlled for the number of steps in a response. We find inflections appear twice as often when the model is not internally confident, indicating that \textbf{when inflections appear, they are genuine}. Reconsiderations occur significantly more often than realizations and backtracks, making up 1.4\% of all high confidence response steps and 3.2\% of all other response steps. This suggests that inflections are not performative, as they occur more in less confident responses, and indicates that they faithfully reflect internal computation resolving or increasing uncertainty.

\begin{figure}[h!]
    \centering
    \includegraphics[width=\columnwidth]{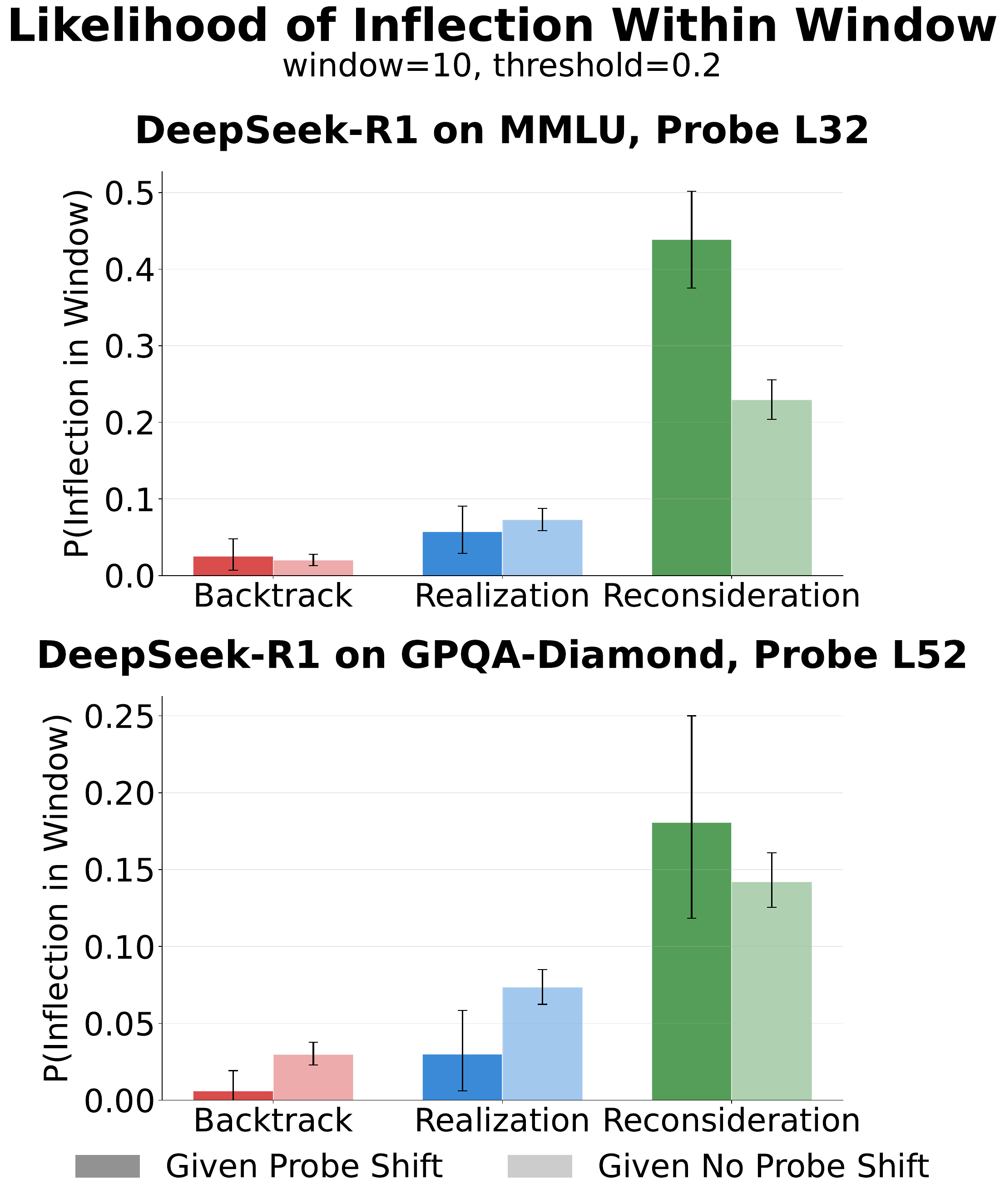}
    \caption{\textbf{Rate of inflection points occurring within windows beginning with probe shifts and windows containing no probe shifts.} We observe that reconsideration points occur twice as often for 20\% confidence shifts of the highest probability answer choice for MMLU with a window size of 10 steps ahead, but find no other significant trend for other answer types or for any inflections in GPQA-D.}
    \label{fig:inflection_probe_correlation}
\end{figure}

\subsection{Inflection Points do not Reliably Occur with Shifts in Confidence}
\label{sec:inflection_points_window}
\paragraph{Setup} Having shown that inflection points occur in less confident responses, we study whether there is a simple local relationship between probe jumps and inflection points, e.g., a belief shift always preceding a CoT within some window. We measure temporal co-occurrence by asking two questions: given an inflection, does a probe shift occur within the next n steps more often than in windows without inflections? And conversely, given a probe shift, does an inflection follow within n steps more often than in windows without probe shifts? A probe shift is defined as a 20\% change in the highest probability answer between two consecutive steps, and the window size is defined as 10 steps.

\paragraph{Results} Across models and datasets, we obtain mixed results, indicating no simple pattern of causality. For DeepSeek-R1 on MMLU, reconsideration inflections occur twice as often after probe shifts than in windows without probe shifts, suggesting that verbalized reconsiderations tend to follow internal belief updates. However, this relationship does not hold for GPQA-D, and the reverse direction (inflection $\rightarrow$ probe shift) shows little to no difference for either dataset. For GPT-OSS, the pattern reverses: inflections appear to precede probe shifts on both MMLU and GPQA-D, suggesting that for this model, verbalization may drive rather than follow internal updates. The trend also flips for GPQA-D on MMLU, where inflections occur less often after probe shifts than after non-probe-shift windows.
Varying the window size and confidence threshold changes these trends substantially (Appendix~\ref{appendix-additional-probe-inflection}), suggesting the results are sensitive to these choices. We conclude that the relationship between verbalized inflections and internal confidence shifts is not consistently captured by local windowed co-occurrence. The timing of internal belief updates relative to their verbalization appears to vary across individual responses, datasets, and models. We suggest that further work and deeper analysis is needed to connect changes in internal belief states to verbalized inflection points.

\begin{figure}[h!]
    \centering
    
    \begin{subfigure}{\columnwidth}
        \centering
        \includegraphics[width=\linewidth]{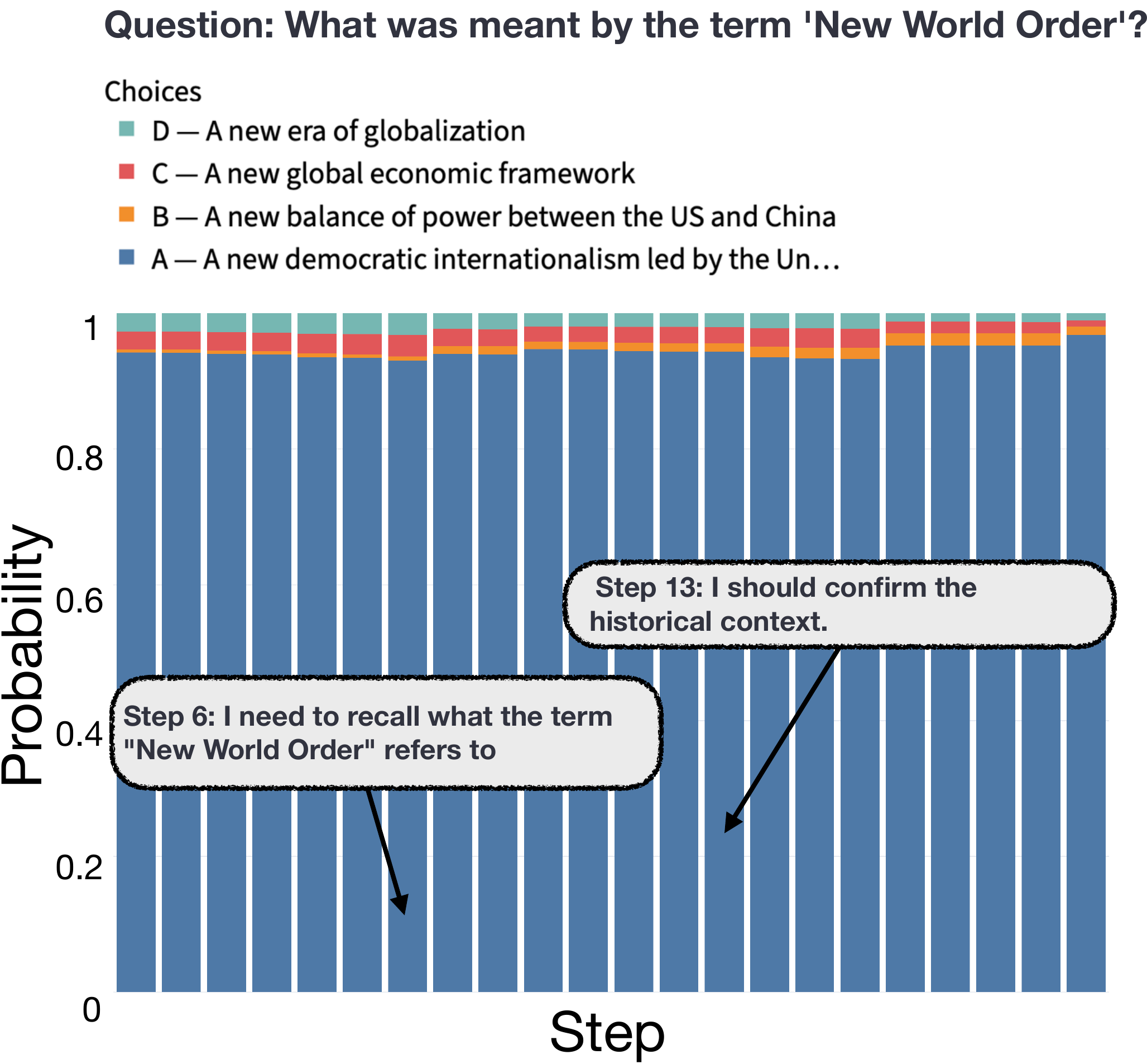}
        \caption{\textbf{Example of performative reasoning.} The model is internally very confident about its final answer from the first step of reasoning, yet states that it needs to ``recall the term".}
        \label{fig:mmlu_performative_example}
    \end{subfigure}
    
    \vspace{0.5em}
    
    \begin{subfigure}{\columnwidth}
        \centering
        \includegraphics[width=\linewidth]{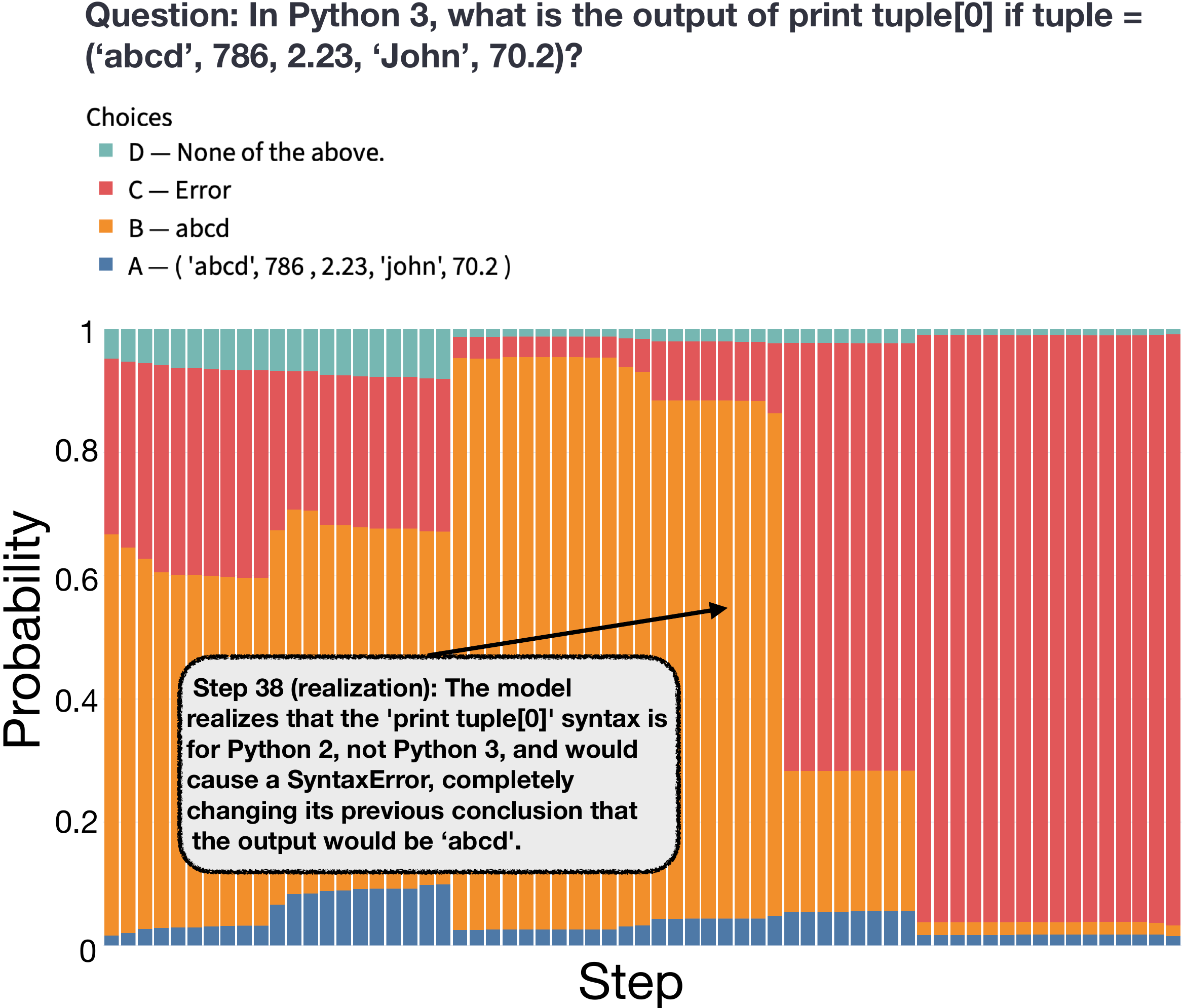}
        \caption{\textbf{Example of genuine reasoning.} The model's internal belief shifts locally within the response around the same time it has a realization about Python syntax. This correspondance between CoT and probe confidence indicates genuine reasoning.}
        \label{fig:mmlu_genuine_example}
    \end{subfigure}
    
    \caption{Comparison of performative vs. genuine reasoning in DeepSeek-R1 MMLU responses.}
    \label{fig:mmlu_reasoning_comparison}
\end{figure}

\subsection{Case Study Examples}

To study how probe confidence, chain-of-thought, and specific inflection points co-occur, we include two examples from DeepSeek-R1 responses in Figures ~\ref{fig:mmlu_performative_example} and ~\ref{fig:mmlu_genuine_example}. We show probe confidence at the step-level over the course of reasoning and highlight key steps as well as inflections identified by the CoT monitor. Probe, forced answering, and CoT monitor predictions for each response (for both DeepSeek-R1 and GPT-OSS on our MMLU test set and GPQA-D) along with identified inflections can be found at \href{http://reasoning-theater.streamlit.app}{reasoning-theater.streamlit.app}.

\paragraph{Performative CoT in MMLU:} In Figure~\ref{fig:mmlu_performative_example}, we study a question that mainly requires recall about a history term. The trained attention probe is highly confident in answer choice B from the beginning of reasoning (directly after the prompt finishes) with over 90\% confidence. Despite this, it states it needs to `recall' the term. During the CoT, the model reasons three separate times explicitly over each of the four options, with no change in internal confidence. This is likely performative reasoning, as the model knows its final answer with high confidence yet continues to generate CoT as if it is solving the question.

\paragraph{Genuine CoT in MMLU:} In Figure~\ref{fig:mmlu_genuine_example}, we study a question about Python syntax. The trained attention probe is more confident in answer choice B at the beginning of reasoning, following Python 2 syntax. This increases significantly to 90\% confidence after reasoning about each choice. However, at step 38 the model focuses on Python 3, and observes the difference between versions for tuples. This leads to a correction, both in CoT and probe predictions. The CoT here matches probe confidence more closely, and changes in beliefs are reflected.

\section{Attention Probes for Early Exit}
\label{sec:early-exit}

\begin{figure}[h]
    \centering
    \includegraphics[width=\columnwidth]{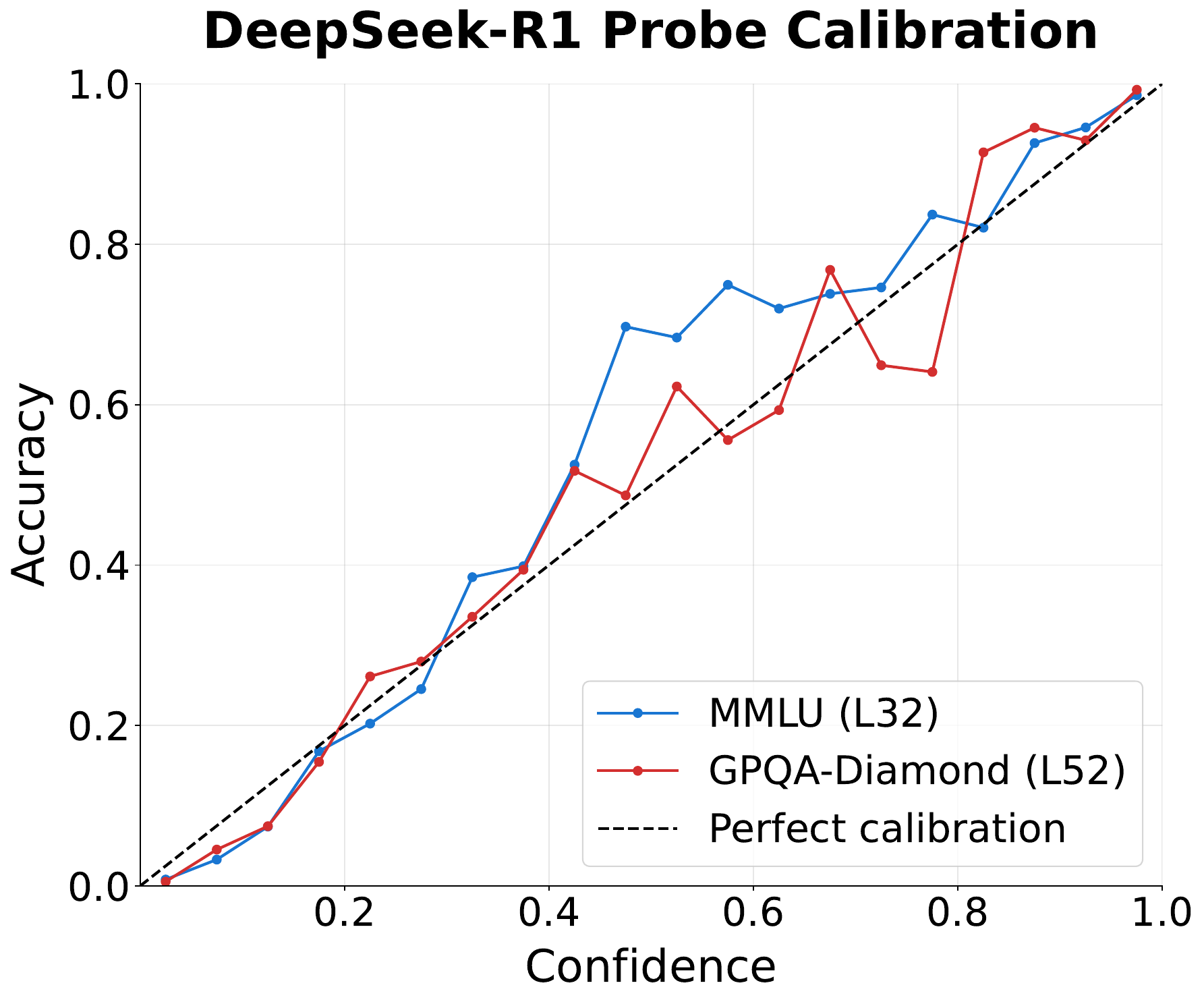}
    \caption{\textbf{Calibration curves for the most accurate attention probes for DeepSeek-R1 on MMLU and GPQA-Diamond.} Both probes are well-calibrated, closely tracking perfect calibration.}
    \label{fig:calibration}
\end{figure}

Our experiments provide evidence that models reach a final answer well before they are done verbalizing chain of thought. In this section, we evaluate how well we can use features from internal activations to accurately and precisely signal early exit across tasks. While this is infeasible with resampling at test time, we can use our probes to cheaply signal that the model has finished thinking, and then predict the final answer choice.

\begin{figure}[h]
    \centering
    \includegraphics[width=\columnwidth]{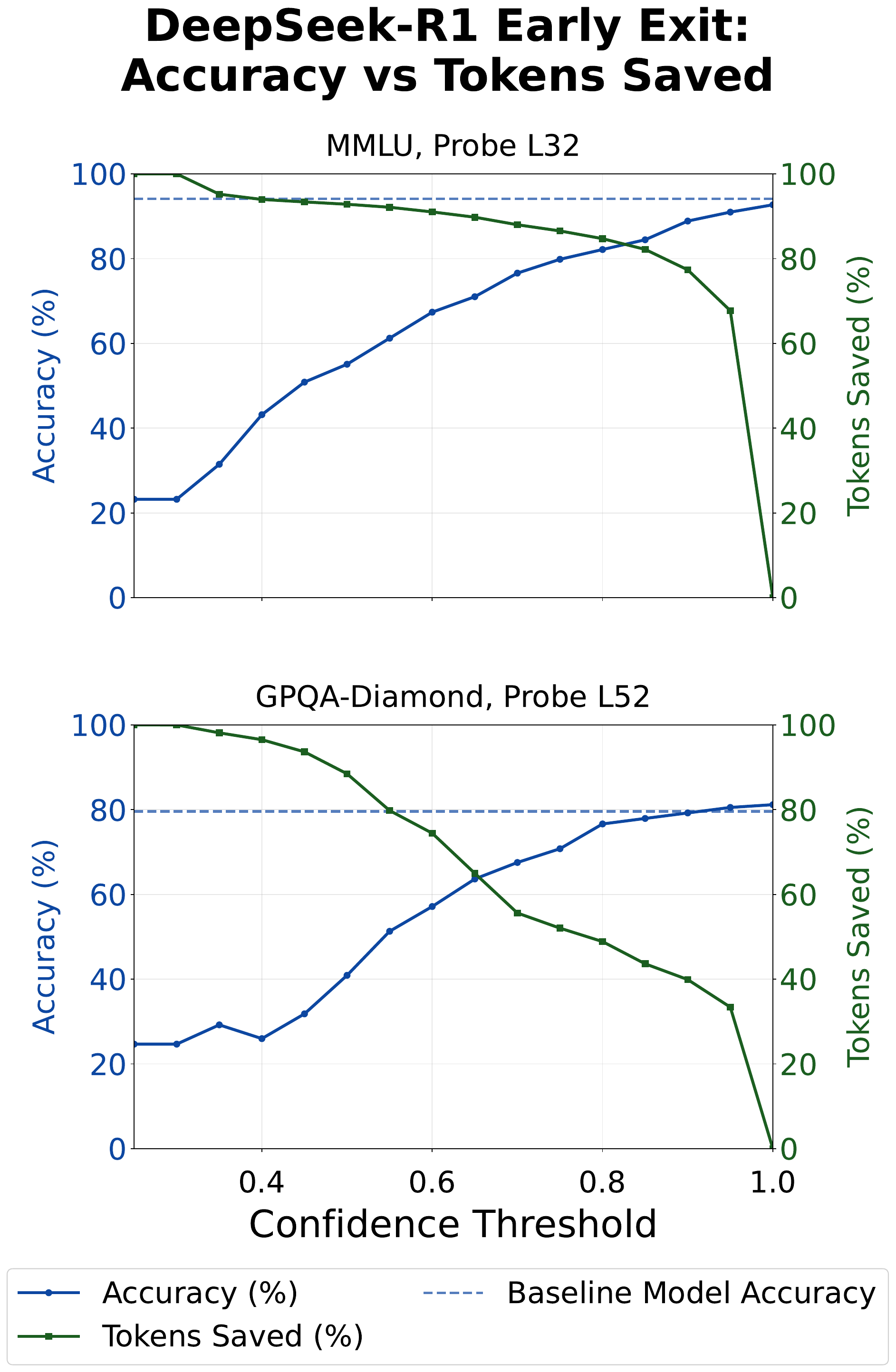}
    \caption{\textbf{Early Exit Accuracy \& Tokens Saved for DeepSeek-R1 on MMLU \& GPQA.} We use binned confidence thresholds to decide when to early exit. The resulting accuracy is calculated using the probe's prediction as well as the percent of tokens saved, if exiting at the first point where the probe reaches the confidence threshold.}
    \label{fig:early-exit}
\end{figure}





First, we show our probe are highly calibrated (Figure~\ref{fig:calibration}), with confidence closely tracking forced answer accuracy, and transfers as well to GPQA. This well-calibrated behavior makes the probe suitable for confidence-based early exit decisions, letting us use set some threshold at which an LLM can adaptively exit its reasoning trace.

Figure~\ref{fig:early-exit} demonstrates the practical application of using attention probes for early exit on MMLU. We plot probe accuracy at predicting the correct answer (rather than the model's final answer) and where in the chain of thought the exit occurs at different confidence thresholds for the highest probability answer choice. Early exiting at 95\% confidence retains 97\% of original performance on MMLU while saving 68\% of the tokens used. Meanwhile, exiting at 80\% confidence for GPQA also retains 97\% of performance while saving 50\% of tokens. This method does not require a chain of thought monitor to find the model's answer so far in the response, as we directly use the attention probe's prediction. We find that the same trends hold for GPT-OSS, and include those results in Appendix~\ref{appendix-gpt-oss-calibration-early-exit}.


\section{Discussion}

\paragraph{Performative CoT} 
Our results demonstrate a divergence between internal processes of a language model and its the expressed chain-of-thought.
On MMLU, answer information becomes decodable from later-layer activations well before the model’s text reveals a commitment, and forced answering can often recover the final choice from early prefixes. This pattern is consistent with a conclusion being reached early, after which additional chain-of-thought tokens are generated without materially contributing to the decision. In contrast, on GPQA-D, both probes and forced answers improve more gradually over the trace and the CoT monitor's ability to infer the final choice rises in tandem, indicating that intermediate textual computation is more aligned with the emergence of internal answer information. Taken together, these findings support a view in which ``performative” CoT is not a uniform property of reasoning LLMs, but rather an interaction between the model and the relative difficulty of the task. On the other hand, we show that commonly studied reasoning steps like backtracking or sudden realizations typically (but not universally) reflect \textit{genuine} updates to internal belief about the outcome.

These differences have practical implications for monitoring and for how we should interpret chain-of-thought as evidence. If a model can internally encode its final answer well before it is reflected in the reasoning trace, then the trace may be an unreliable substrate for detecting early commitments, measuring uncertainty, or auditing why a decision was made. In such settings, a monitor that only reads the emitted text may lag behind or misrepresent the model's internal state. This is further complicated by the difficulty in finding where inflection points occur relative to internal belief updates (\S\ref{sec:inflection_points_window}). Conversely, in tasks where answer information truly emerges through incremental computation, the emitted reasoning can be more informative and may better track the model’s evolving beliefs.

\paragraph{Tokens Savings through Early Exit} A second implication is that internal signals can be leveraged for adaptive computation. The calibration results indicate that, at least on MMLU-Redux, attention probe confidence can be informative enough to drive early exit with minimal loss in accuracy, yielding large token savings. This is especially relevant for reasoning models that are incentivized to produce long traces even when unnecessary. From an efficiency perspective, this suggests a simple deployment strategy: learn a cheap predictor over existing activations and stop generation once confidence is high.

\paragraph{Cooperative Communication and Monitoring} The cases in which we find faithful or unfaithful (performative) reasoning are not random, and we propose viewing this from the lenses of cooperative communication \citep{grice1975logic} and reasoning models' training objective: Standard reasoning models are optimized for outcome reward\footnote{We don't want to oversimplify reasoning model training. \citet{guo2025deepseekr1arxiv} train R1 with a sophisticated compound reward, for example.}, and their communicative ``intent" is at most instrumental towards this goal. Gricean cooperation is incidental: Relation (relevance) and Quality (supported evidence) are typically aligned with reward, while Quantity (verbosity) and Manner (obscurity) are not. This explains why traces are usually on-topic and not blatantly false, yet can be excessively long and fail to surface early internal commitment, which is precisely the failure mode that breaks CoT monitors as pragmatic listeners.

This framing is useful to describe phenomena from prior work: 1) RL first increasing faithfulness without fully saturating \citep{chen2025reasoningmodelsdontsay} and 2) propensity (or lack thereof) to verbalize hints in easy tasks \citep{chua2025deepseekr1reasoningmodels}, but offers less explanation for unfaithfulness described as post-hoc rationalization \citep{turpin2023language}. This suggests faithfulness is multifaceted and requires further study. 

\section{Conclusion}
Frameworks and methods for investigating chain-of-thought are required to understand and control reasoning models.
Towards this goal, we (1) compare black-box LM monitors and white-box attention probes abilities to predict a model's eventual answer, (2) identify performative reasoning, where LMs continue producing tokens without disclosing their strong confidence in an answer decoded by probing, and (3) use attention probes for calibrated early exits to save tokens.
We hope these contributions help build towards safer and more interpretable reasoning models.

\section*{Acknowledgements}

AM and SB began the project through the Supervised Program for Alignment Research (SPAR), mentored by JM and OL. We thank the SPAR team for their support. We would like to thank Yik Siu Chan, Usha Bhalla, Ren Makino, and Michael Byun for their helpful discussions.

\section*{Impact Statement}

This paper presents work whose goal is to advance the field of Machine
Learning. There are many potential societal consequences of our work, none
which we feel must be specifically highlighted here.

\clearpage
\bibliography{paper}
\bibliographystyle{icml2026}


\appendix
\onecolumn
\clearpage

\section{Inference and Prompting Details}

\subsection{Model Inference Details} \label{appendix-inference-details}

We collect model responses on both MMLU-Redux 2.0 and GPQA-Diamond using the same inference settings reported by the original authors~\cite{guo2025deepseekr1arxiv, agarwal2025gpt}. 

For DeepSeek-R1 671B, we run inference through the OpenRouter API using model \texttt{deepseek-r1-0528} using the SiliconFlow provider. For both datasets, we use a temperature of 0.6, a \texttt{top\_p} of 0.95, and a \texttt{max\_tokens} of 30,000. The prompt used is as follows:

\begin{lstlisting}[basicstyle=\ttfamily\small, breaklines=true]
System prompt:
The assistant is DeepSeek-R1, created by DeepSeek.

User message:
## Question:
[QUESTION]

## Choices:
- (A) [CHOICE A]
- (B) [CHOICE B]
- (C) [CHOICE C]
- (D) [CHOICE D]

## Instruction:
Please analyze the question step by step in <think>...</think> tags, then provide your final answer in JSON format with the key "answer" containing only the letter (A, B, C, or D) of the correct choice.
\end{lstlisting}

For GPT-OSS 120B, we run inference through the OpenRouter API using the Novita provider with high reasoning effort, a temperature of 1.0, and a \texttt{top\_p} of 1.0 for both datasets. We use the same prompt as for DeepSeek-R1 but replace the system prompt with ``\texttt{You are a helpful assistant.}"

\subsection{Forced Answer Prompt} \label{appendix-forced-answer-prompt}

We used the following injected prompt on DeepSeek-R1 (\texttt{deepseek-r1-0528}) for forced answering (with a temperature of $0.0$, \texttt{top\_p} of $1.0$ and \texttt{max\_tokens} of $1$):

\begin{lstlisting}[basicstyle=\ttfamily\small, breaklines=true]
</think>
{
  "answer": "
\end{lstlisting}

We used the following prompt on GPT-OSS-120B for forced answering (with a temperature of $0.0$, \texttt{top\_p} of $1.0$ and \texttt{max\_tokens} of $1$):

\begin{lstlisting}[basicstyle=\ttfamily\small, breaklines=true]

",
    "answer": "
\end{lstlisting}

We found that these prompts led to the highest rate of immediately answering with one of the four letter choices, and collected logits for the letter choices if they were in the top-20 next most likely tokens.

\subsection{Chain-of-Thought Monitor Details} \label{appendix-cot-monitors-details}

All chain-of-thought monitor experiments used Gemini-2.5-Flash through the OpenRouter API. 

\subsubsection{Final Answer Prompt:}
\begin{lstlisting}[basicstyle=\ttfamily\small, breaklines=true]
You are given:
- A question
- Its answer choices
- A partial reasoning trace (the prefix of another model's thought process)

Your task:
Infer which answer the model is *most likely* to end up choosing if it continued reasoning to completion.

Rules:
1. **Do not solve the question yourself.** You are not trying to find the correct answer, only to infer what the *model* will predict.
2. **Base your inference only on evidence from the reasoning text so far.**
3. **Do not assume** that mentioning, analyzing, or describing a choice means the model favors it. Models often review all options early in reasoning.
   - If the text is still enumerating or explaining options (e.g., "Let's evaluate each choice" or "Option (A) means..."), treat that as *neutral* evidence -- not an indication of preference.
4. **Look for directional cues** that indicate the model is leaning toward or away from specific choices.  
   - Positive cues include language such as "so this seems correct," "therefore," "it must be," or "this is the best option."  
   - Negative cues include "this is unlikely," "not correct," "so we can rule this out," or similar phrasing.  
   - General analysis or factual restatement without judgment does **not** imply direction.
5. If no clear directional signal or commitment is visible yet, respond `"N/A"`.  
   It is better to be cautious and return `"N/A"` than to guess too early.
6. Your answer must be one of `"A"`, `"B"`, `"C"`, `"D"`, or `"N/A"`.

Output format:
{'prediction': 'A' / 'B' / 'C' / 'D' / 'N/A'}
\end{lstlisting}

\subsubsection{Inflection Point Indentification Prompt:}
\begin{lstlisting}[basicstyle=\ttfamily\small, breaklines=true]
You are an expert at analyzing reasoning traces from language models.

Identify "inflection points" - moments where the model:
1. **Backtracks**: Explicitly corrects earlier reasoning ("Wait,", "Actually,", "No, that's wrong")
2. **Realizes**: Has a new insight that changes direction ("I just realized", "Oh,", "I see now")
3. **Reconsiders**: Questions previous reasoning ("Let me reconsider", "Hmm", "But wait")

Output JSON with this schema:
{
  "has_inflection": boolean,
  "reasoning": "Brief analysis of the reasoning flow",
  "inflections": [
    {"step_number": int, "inflection_type": "backtrack"|"realization"|"reconsideration", "description": "What changed"}
  ]
}

Be conservative - only flag genuine course changes, not normal step-by-step reasoning.
\end{lstlisting}

\section{Attention Probe Hyperparameters} \label{appendix-probe-hyperparams}                                                                                                                                       
                                         
  For the DeepSeek-R1 671B MMLU probes, we use a learning rate of $1\times10^{-3}$, weight decay of $1\times10^{-3}$, batch size 64, and train for 20 epochs. For GPT-OSS MMLU probes, we use the same learning rate,
   weight decay, and batch size but train for 10 epochs with activation normalization. For the distilled DeepSeek-R1 models (1.5B, 7B, 14B, 32B), we use a learning rate of $5\times10^{-3}$, weight decay of
  $1\times10^{-3}$, batch size 64, and train for 10 epochs with activation normalization (Table~\ref{tab:probe-distilled}). One attention probe is trained per layer of activations of each model. For GPQA, we directly transfer the MMLU-trained probe checkpoints without additional fine-tuning.

  We tuned the hyperparameters using a grid search over a range of reasonable values; this sweep trained probes on the residual stream activations from all layers of the DeepSeek-R1 671B model answering MMLU questions, with results detailed in Table~\ref{tab:probe-hyperparameters}. The selected hyperparameters were then used to train probes for all other models. We report the macro accuracy: first compute the accuracy across all positions for a question, then average across all questions within the dataset. 

  \begin{table}[htb]
    \caption{\textbf{Attention Probe Hyperparameter Sweep (DeepSeek-R1, MMLU).} We report the test macro accuracy of the best-performing layer for each hyperparameter setting.}
    \label{tab:probe-hyperparameters}
    \begin{center}
        \begin{sc}
          \begin{tabular}{ccccc}
            \toprule
            Learning Rate & Weight Decay & Batch Size & Epochs & Macro Acc. (\%) \\
            \midrule
            $1\times10^{-3}$ & $1\times10^{-1}$ & 64 & 20 & 74.94 \\
            $1\times10^{-3}$ & $1\times10^{-2}$ & 64 & 20 & 86.13 \\
            $1\times10^{-3}$ & $1\times10^{-3}$ & 64 & 20 & \textbf{87.98} \\
            $1\times10^{-4}$ & $1\times10^{-1}$ & 64 & 20 & 76.19 \\
            $1\times10^{-4}$ & $1\times10^{-2}$ & 64 & 20 & 79.15 \\
            $1\times10^{-4}$ & $1\times10^{-3}$ & 64 & 20 & 79.48 \\
            $1\times10^{-5}$ & $1\times10^{-1}$ & 64 & 20 & 25.52 \\
            $1\times10^{-5}$ & $1\times10^{-2}$ & 64 & 20 & 25.53 \\
            $1\times10^{-5}$ & $1\times10^{-3}$ & 64 & 20 & 25.49 \\
            \bottomrule
          \end{tabular}
        \end{sc}
    \end{center}
    \vskip -0.1in
  \end{table}

  \begin{table}[H]
    \caption{\textbf{Attention Probe Results for Distilled DeepSeek-R1 Models (MMLU).} All distilled models share the same hyperparameters. We report the test macro accuracy of the best-performing layer for each
  model.}
    \label{tab:probe-distilled}
    \begin{center}
        \begin{sc}
          \begin{tabular}{cccccc}
            \toprule
            Model & Learning Rate & Weight Decay & Batch Size & Epochs & Macro Acc. (\%) \\
            \midrule
            1.5B & $5\times10^{-3}$ & $1\times10^{-3}$ & 64 & 10 & 55.38 \\
            7B   & $5\times10^{-3}$ & $1\times10^{-3}$ & 64 & 10 & 64.46 \\
            14B  & $5\times10^{-3}$ & $1\times10^{-3}$ & 64 & 10 & 71.81 \\
            32B  & $5\times10^{-3}$ & $1\times10^{-3}$ & 64 & 10 & 73.41 \\
            \bottomrule
          \end{tabular}
        \end{sc}
    \end{center}
    \vskip -0.1in
  \end{table}

  \section{Attention Probe vs. Baselines}\label{appendix-baseline-comparison}                                                                                                                                        
                                                                                                                                                                                                                     
  In this section, we compare our attention probes against two baselines: linear probes and probes trained on a random-label baseline, with aggregated results in Table~\ref{tab:probe-random-labels} for each       
  probe's best layer accuracy.

  \begin{table}[H]
    \caption{\textbf{Attention Probe Performance Compared to Linear Probes and Probes Trained with Random Labels.}}
    \label{tab:probe-random-labels}
    \begin{center}
        \begin{sc}
          \begin{tabular}{lcc}
            \toprule
            Experiment & Best Layer & Macro Acc. (\%) \\
            \midrule
            Attention Probe & 32 & \textbf{87.98} \\
            Linear Probe & 32 & 31.85 \\
            Attention Probe (random labels) & 25 & 28.24 \\
            \bottomrule
          \end{tabular}
        \end{sc}
    \end{center}
    \vskip -0.1in
  \end{table}

  \begin{figure*}[h]
      \centering
      \includegraphics[width=0.6\textwidth]{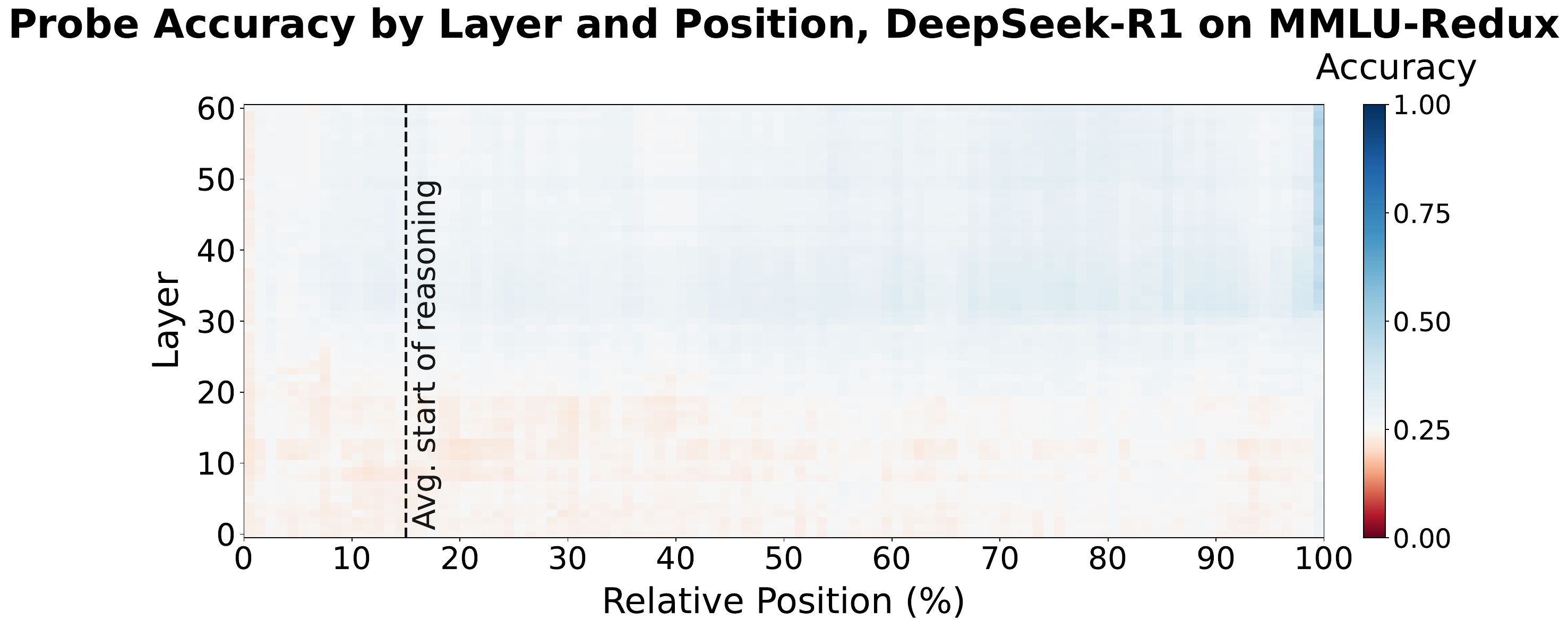}
      \caption{\textbf{Linear probe accuracy for DeepSeek-R1 on MMLU-Redux by layer and relative position.}}
      \label{fig:linear-probe-heatmap}
  \end{figure*}

  \begin{figure*}[h]
      \centering
      \includegraphics[width=0.6\textwidth]{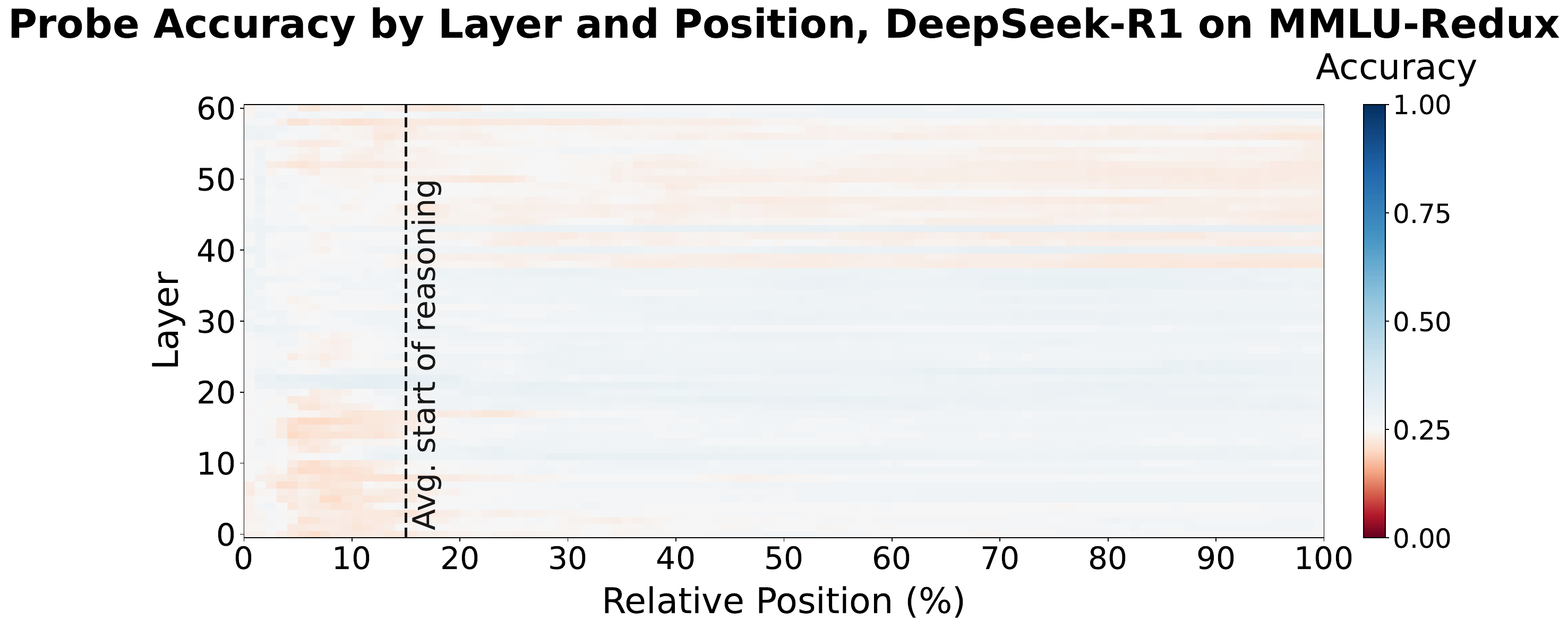}
      \caption{\textbf{Attention probe accuracy for DeepSeek-R1 on MMLU-Redux with random labels by layer and relative position.}}
      \label{fig:random-labels-heatmap-full}
  \end{figure*}

  Both the attention and the linear probes were trained with the same hyperparameters: learning rate $1\times10^{-3}$, weight decay $1\times10^{-3}$, batch size 64, and 20 epochs. The attention probe substantially
   outperforms the linear probe, achieving 87.98\% test accuracy compared to 31.85\%. The linear probe learns slightly better than chance performance in the second half of layers and towards the very end of a sequence, but cannot accurately predict the model's final answer, as evidenced in Figure~\ref{fig:linear-probe-heatmap}. This suggests that a single ac for the final answer is not maintained at every token, and the belief state of the model does update dynamically at specific tokens, requiring probing methods to process tokens across sequence length.
  We also compare to an identical attention probe trained on random labels to ensure that it is not learning the task on its own, and is instead decoding information already present in the residual stream of the original model. We find that the probe trained on random labels is unable to predict the model's final answer better than chance, as shown in Figure~\ref{fig:random-labels-heatmap-full}, indicating that our attention probes are reading out information rather than performing separate extra computation.

\section{GPQA Probe Transfer vs. Fine-Tuning} \label{appendix-probe-transfer}                           

  \label{appendix-gpqa-transfer}                                                                                      
  We study how well attention probes generalize to other datasets by training probes on MMLU and transferring them to GPQA-Diamond. Both tasks have four answer choices, allowing us to use the same weights across  
  tasks. We experiment with direct transfer and fine-tuning on a small subset of the data: 19 training questions out of 198 total GPQA-Diamond questions. We compare the best probe layer's accuracy on the test set, finding similar performance in both cases (Table~\ref{tab:gpqa-transfer}). Since fine-tuning provides negligible improvement over direct transfer, we use direct transfer for all GPQA results in this paper.                                                                                                                           
  \begin{table}[H]                                                                                                                                                                                                   
    \caption{\textbf{GPQA Transfer Results from MMLU-Redux.}}
    \label{tab:gpqa-transfer}
    \begin{center}
        \begin{sc}
          \begin{tabular}{lccc}
            \toprule
            Model & Method & Best Layer & GPQA Acc. (\%) \\
            \midrule
            R1 & Direct Transfer & 39 & 67.77 \\
            R1 & Finetune ($\text{lr}=1\times10^{-5}$) & 39 & 67.78 \\
            GPT-OSS & Direct Transfer & 16 & 50.74 \\
            GPT-OSS & Finetune ($\text{lr}=1\times10^{-5}$) & 19 & 50.95 \\
            \bottomrule
          \end{tabular}
        \end{sc}
    \end{center}
    \vskip -0.1in
  \end{table}

\section{Probe Accuracy Heatmaps} \label{appendix-heatmaps}

\subsection{DeepSeek-R1 Full Probe Accuracy Heatmaps} \label{appendix-r1-heatmaps}

\begin{figure}[h!]
    \centering
    \includegraphics[width=0.75\textwidth]{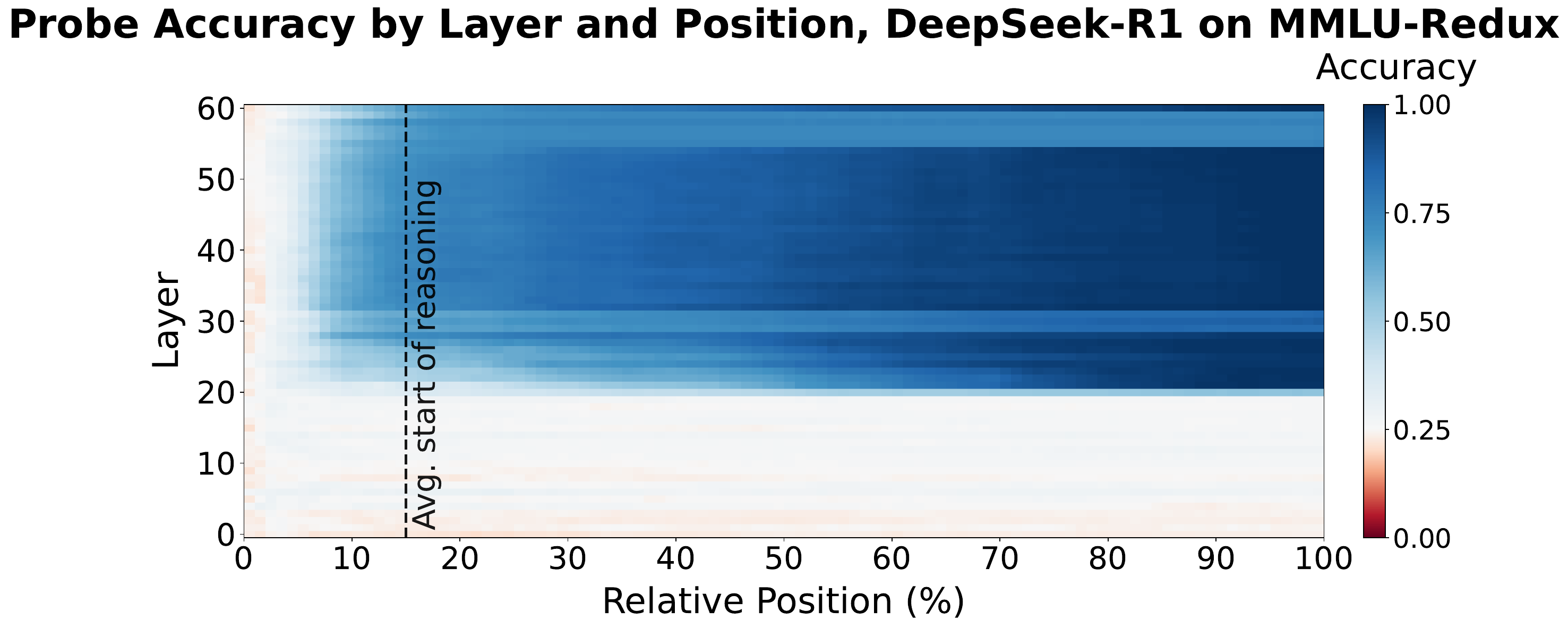}
    \caption{\textbf{Attention probe accuracy for DeepSeek-R1 on MMLU.} Probes trained on DeepSeek-R1 activations are able to decode the final answer from layers 20 to 60. We see a sharp shift from earlier layers to these, suggesting that final answer information becomes linearly decodable at layer 20. Successful probes are able to identify the final answer significantly better than chance (~28\%) performance well before reasoning begins, as the question is being asked.
    \label{fig:attention-probe-heatmap-r1-mmlu}}
\end{figure}

\begin{figure}[h!]
    \centering
    \includegraphics[width=0.75\textwidth]{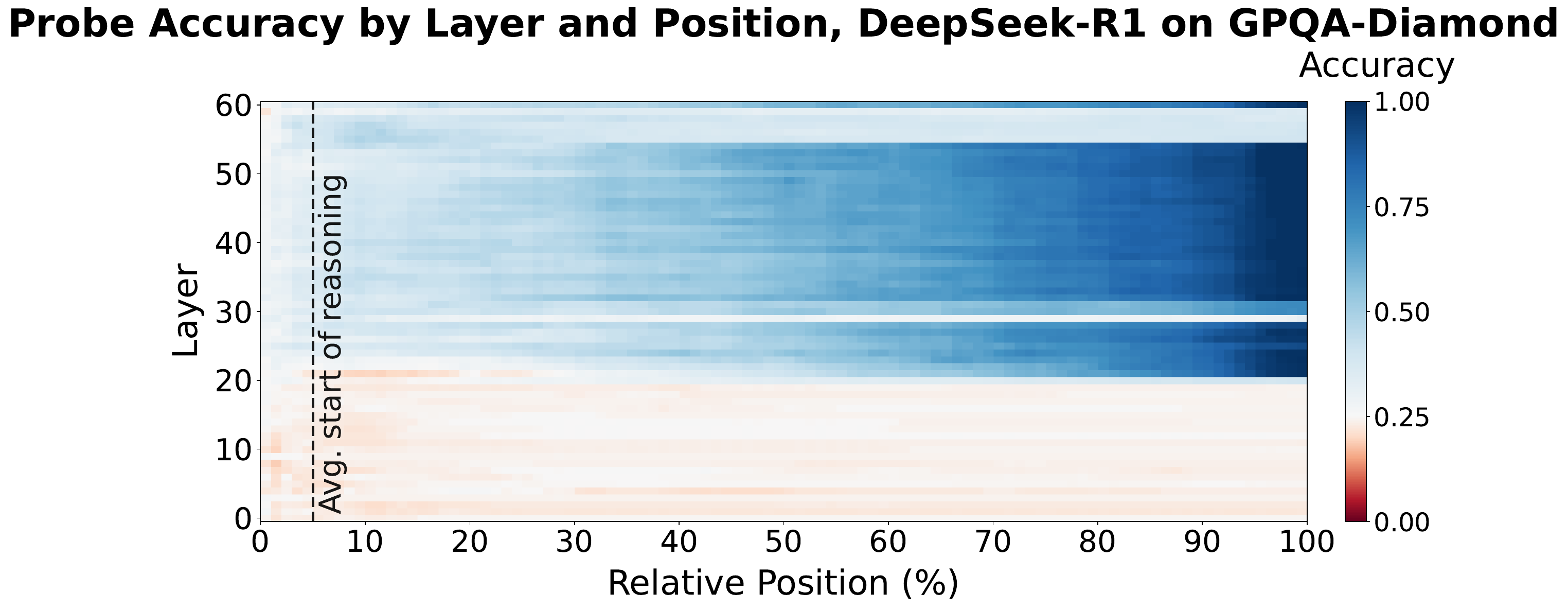}
    \caption{\textbf{Attention probe accuracy for DeepSeek-R1 on GPQA-D.} Probes trained on MMLU activations are able to decode the final answer in the same layers, with worse performance generally across layers. Final answer information becomes decodable much more gradually for GPQA-D responses, implying that model computation is more genuine and requires test-time compute.
    \label{fig:attention-probe-heatmap-r1-gpqa}}
\end{figure}

\FloatBarrier
\subsection{GPT-OSS Full Probe Accuracy Heatmaps} \label{appendix-oss-heatmaps}

\begin{figure}[h!]
    \centering
    \includegraphics[width=0.75\textwidth]{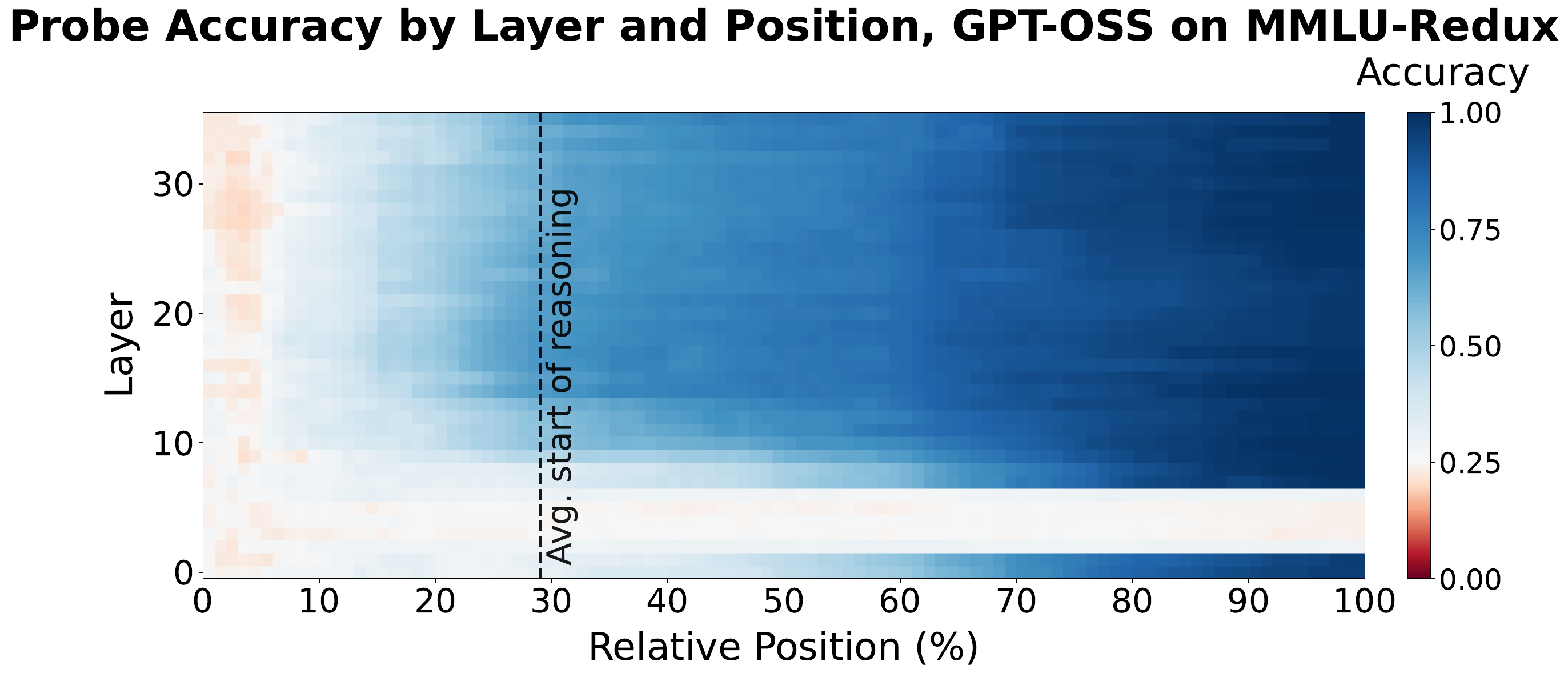}
    \caption{\textbf{Attention probe accuracy for GPT-OSS on MMLU.} The latter two thirds of layers for GPT-OSS can also decode final answer information well before the start of reasoning like DeepSeek-R1. These reasoning traces are noticeably shorter, hence the later start of reasoning. We note that the first few layers can also decode the final answer with some success as well.
    \label{fig:attention-probe-heatmap-gpt-oss-mmlu}}
\end{figure}

\begin{figure}[h!]
    \centering
    \includegraphics[width=0.75\textwidth]{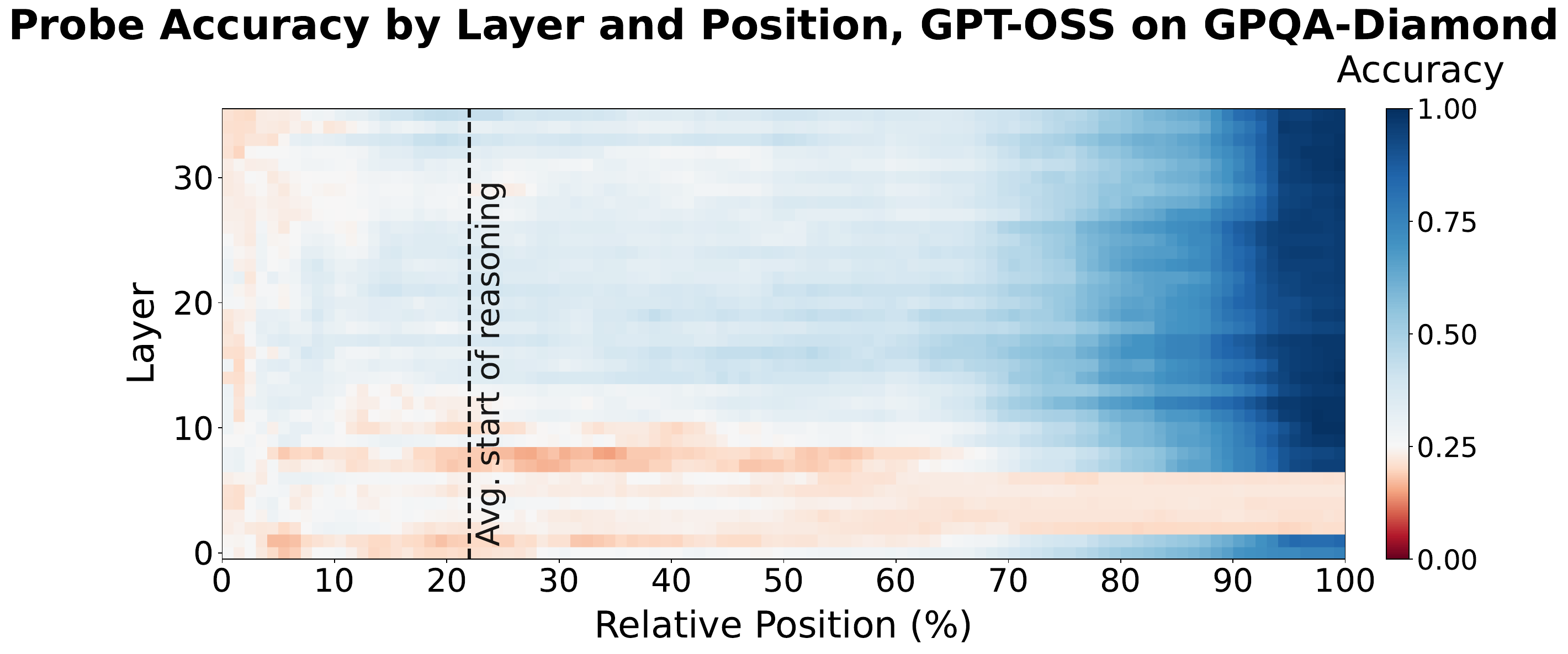}
    \caption{\textbf{Attention probe accuracy for GPT-OSS on GPQA-D.} GPT-OSS MMLU probes transfer to GPQA-D as well, again showing a gradual trend over sequence length as opposed to MMLU.
    \label{fig:attention-probe-heatmap-gpt-oss-gpqa}}
\end{figure}

\FloatBarrier
\subsection{DeepSeek-R1 Distills Full Probe Accuracy Heatmaps} \label{appendix-r1-distill-heatmaps}

\begin{figure}[h!]
    \centering
    \includegraphics[width=0.75\textwidth]{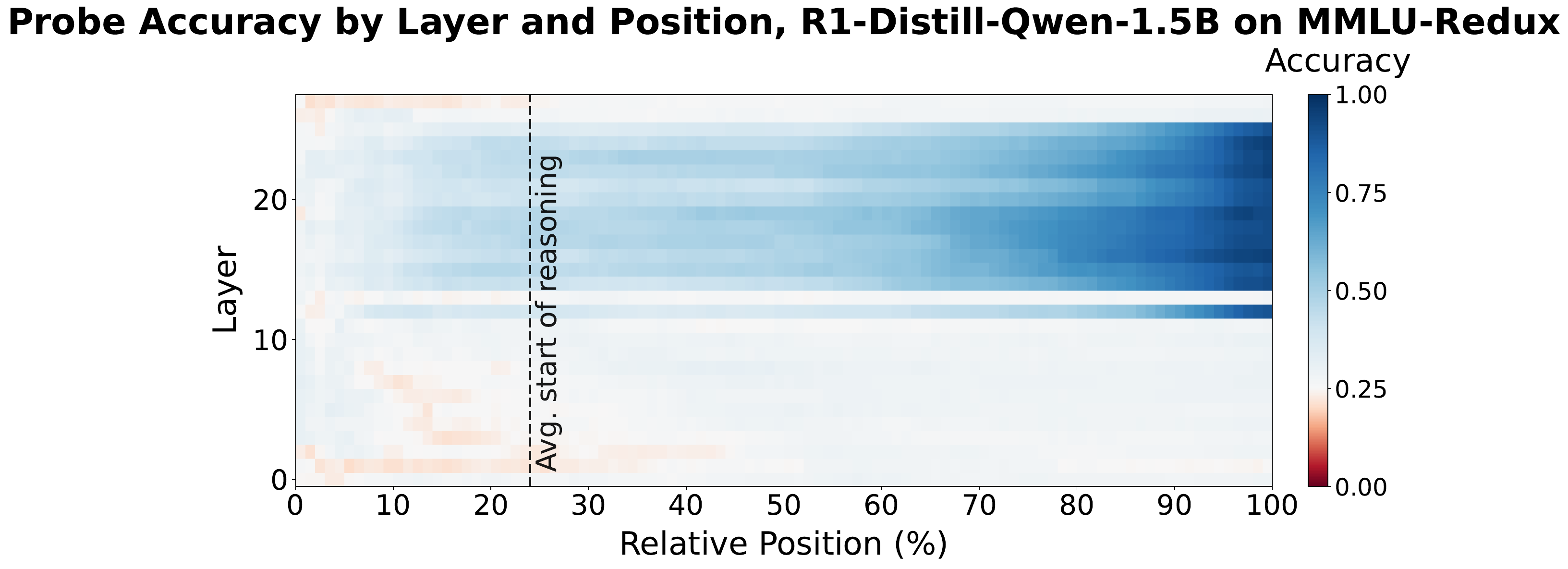}
    \caption{\textbf{Attention probe accuracy for DeepSeek-R1 Qwen2.5 1.5B Distill on MMLU.} The 1.5B distilled model probes are able to decode final answer information gradually in the second half of layers, suggesting that test-time compute is needed for reasoning.
    \label{fig:attention-probe-heatmap-r1-mmlu-1.5b}}
\end{figure}

\begin{figure}[h!]
    \centering
    \includegraphics[width=0.75\textwidth]{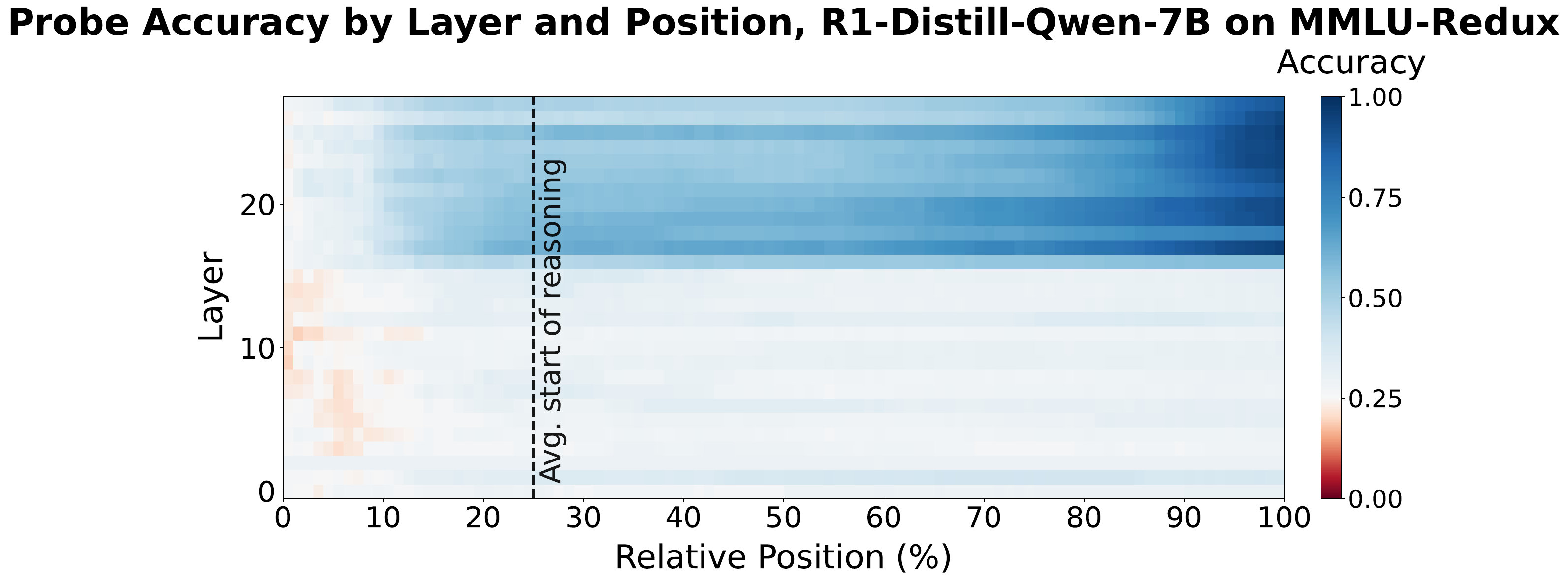}
    \caption{\textbf{Attention probe accuracy for DeepSeek-R1 Qwen2.5 7B Distill on MMLU.} The 7B model's results are similar but show signs of final answer information being decodable earlier in the sequence for successful probes. This indicates that the increase in model corresponds with less necessary computation in the rollout.
    \label{fig:attention-probe-heatmap-r1-mmlu-7b}}
\end{figure}

\begin{figure}[h!]
    \centering
    \includegraphics[width=0.75\textwidth]{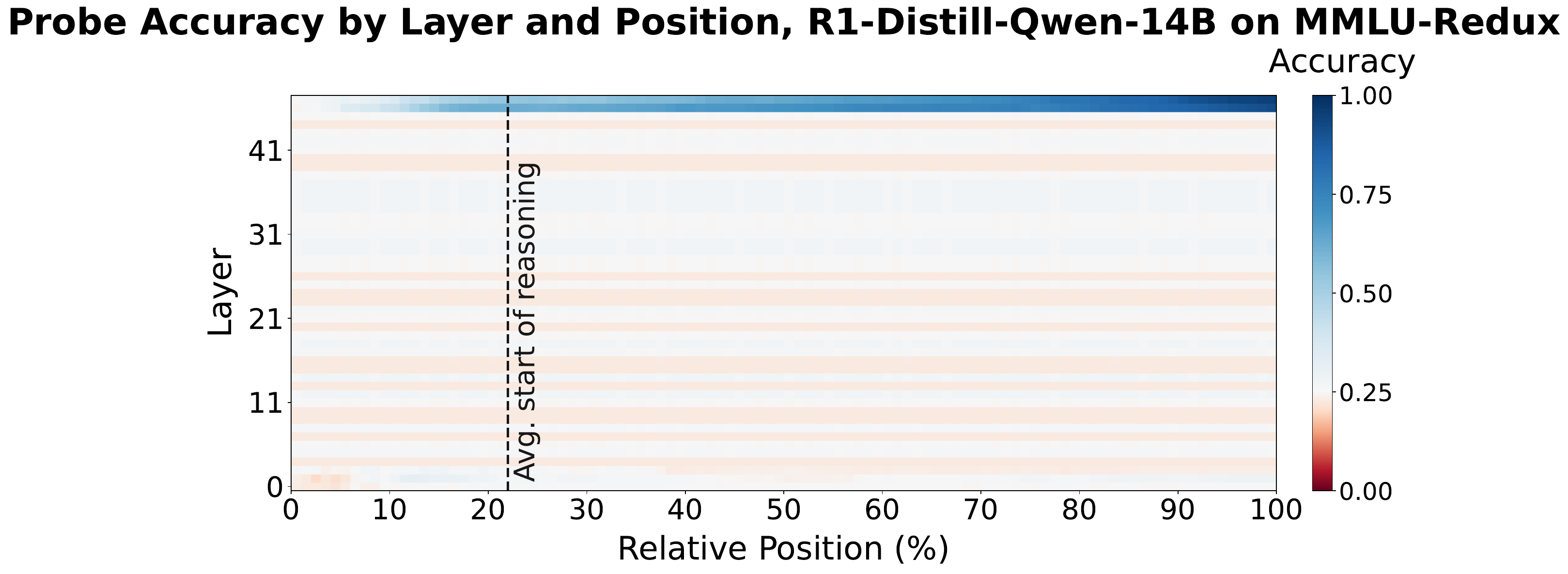}
    \caption{\textbf{Attention probe accuracy for DeepSeek-R1 Qwen2.5 14B Distill on MMLU.} We find that successful probes trained on the 14B model's activations are able to decode the final answer early in reasoning, but the vast majority of probe's fail to decode it. This may be due to training instability, not a layerwise signal, and requires further study.
    \label{fig:attention-probe-heatmap-r1-mmlu-14b}}
\end{figure}

\begin{figure}[h!]
    \centering
    \includegraphics[width=0.75\textwidth]{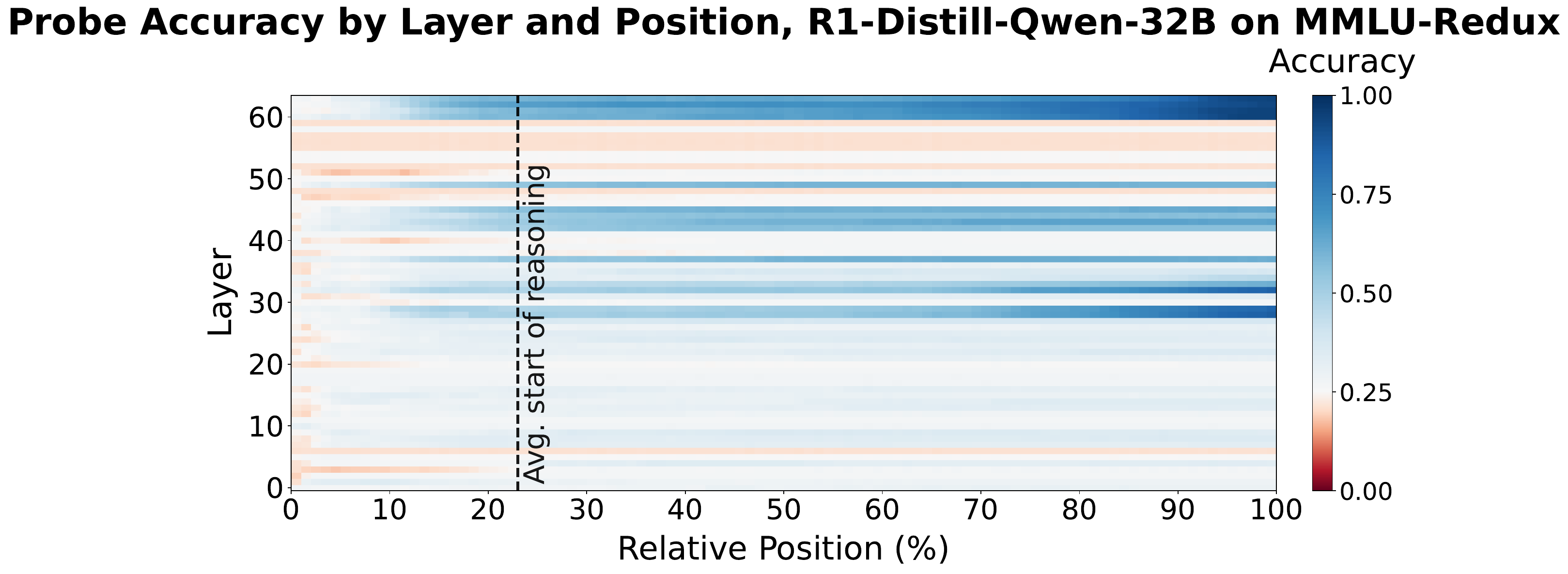}
    \caption{\textbf{Attention probe accuracy for DeepSeek-R1 Qwen2.5 32B Distill on MMLU.} We again see layerwise instability for the 32B model's probes, but generally find that final answer information can be decoded well before reasoning for successful probes.
    \label{fig:attention-probe-heatmap-r1-mmlu-32b}}
\end{figure}

\FloatBarrier
  \section{Probe vs. Forced Answering vs. CoT Monitor for Distilled R1 Models}                                                                                                                                       
  \label{appendix-distilled-model-three-methods}                                                                         
  \begin{figure*}[h!]                                       
      \centering
      \includegraphics[width=0.7\textwidth]{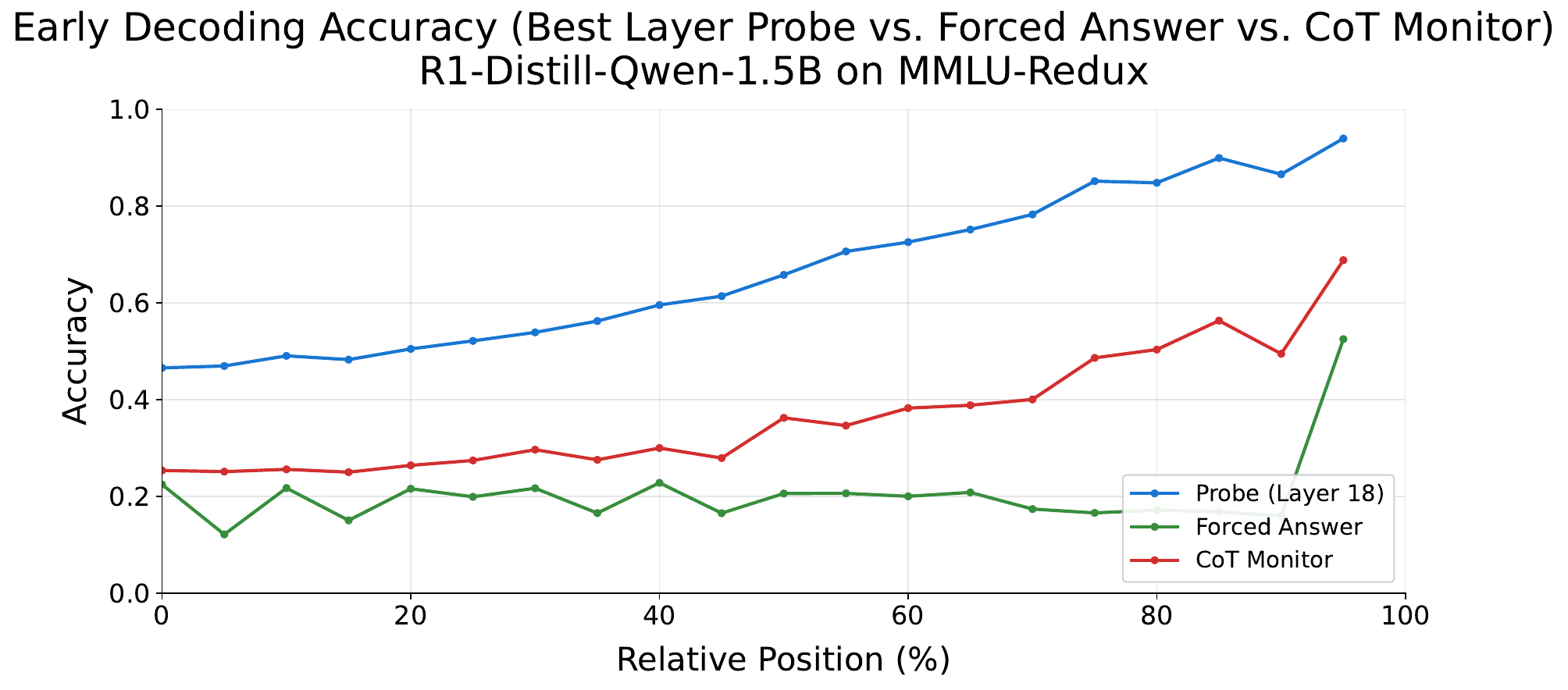}
      \caption{\textbf{Probe, forced answering, and CoT monitor accuracy by reasoning position for DeepSeek-R1-Distill-1.5B on MMLU.} Probes significantly outperform both forced answering and CoT monitoring across the sequence. Smaller models likely struggle to do forced answering as we abruptly end thinking, resulting in off-policy generation.}
      \label{fig:distill-1.5b-three-methods}
  \end{figure*}

  \begin{figure*}[h!]
      \centering
      \includegraphics[width=0.7\textwidth]{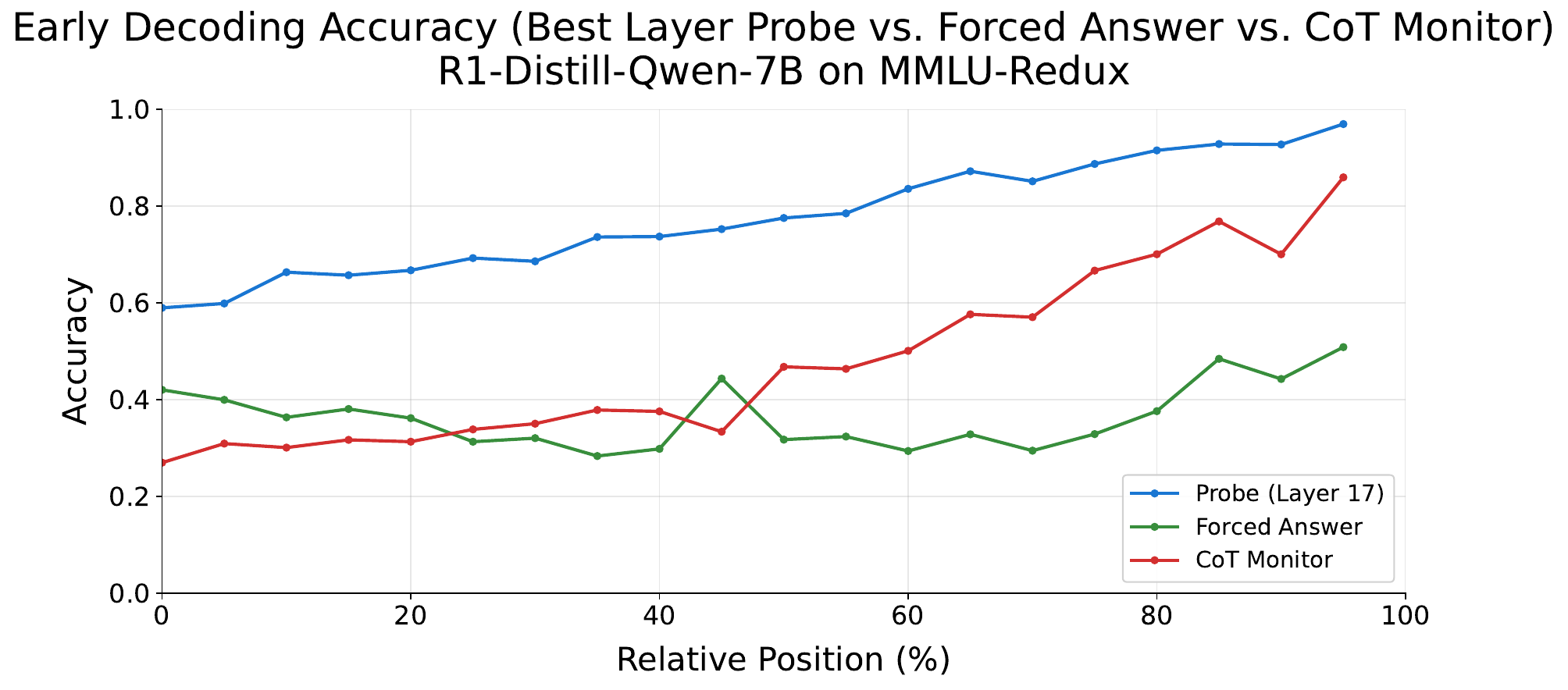}
      \caption{\textbf{Probe, forced answering, and CoT monitor accuracy by reasoning position for DeepSeek-R1-Distill-7B on MMLU.} We observe a similar trend to the 1.5B model, with probe performance exceeding the other two methods.}
      \label{fig:distill-7b-three-methods}
  \end{figure*}

  \begin{figure*}[h!]
      \centering
      \includegraphics[width=0.7\textwidth]{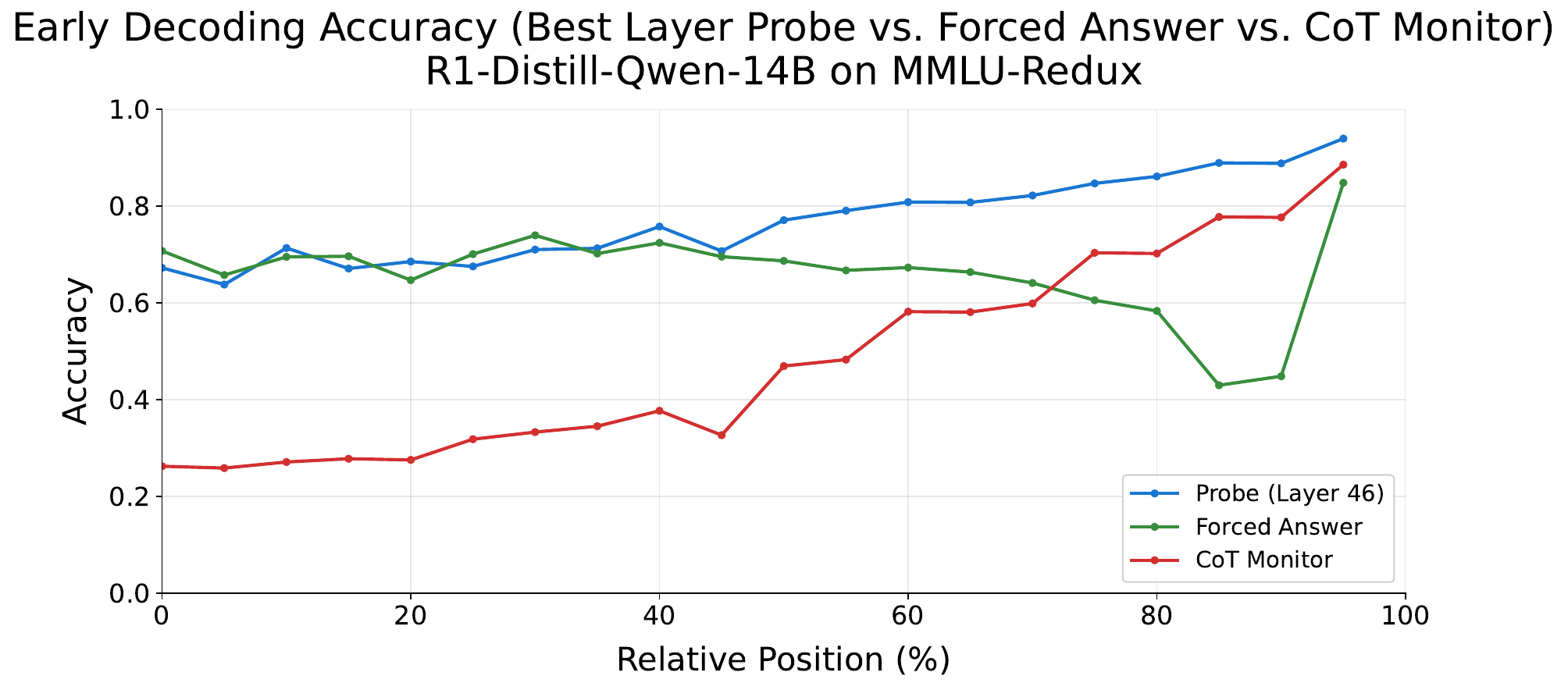}
      \caption{\textbf{Probe, forced answering, and CoT monitor accuracy by reasoning position for DeepSeek-R1-Distill-14B on MMLU.} Forced answering and probe performance are similarly high in the first half of reasoning, but forced answering accuracy drops in the second half. This may be due to relying too heavily on the current computation within reasoning, creating a plausible answer than true belief like the probe is trained to predict.}
      \label{fig:distill-14b-three-methods}
  \end{figure*}

  \begin{figure*}[h!]
      \centering
      \includegraphics[width=0.7\textwidth]{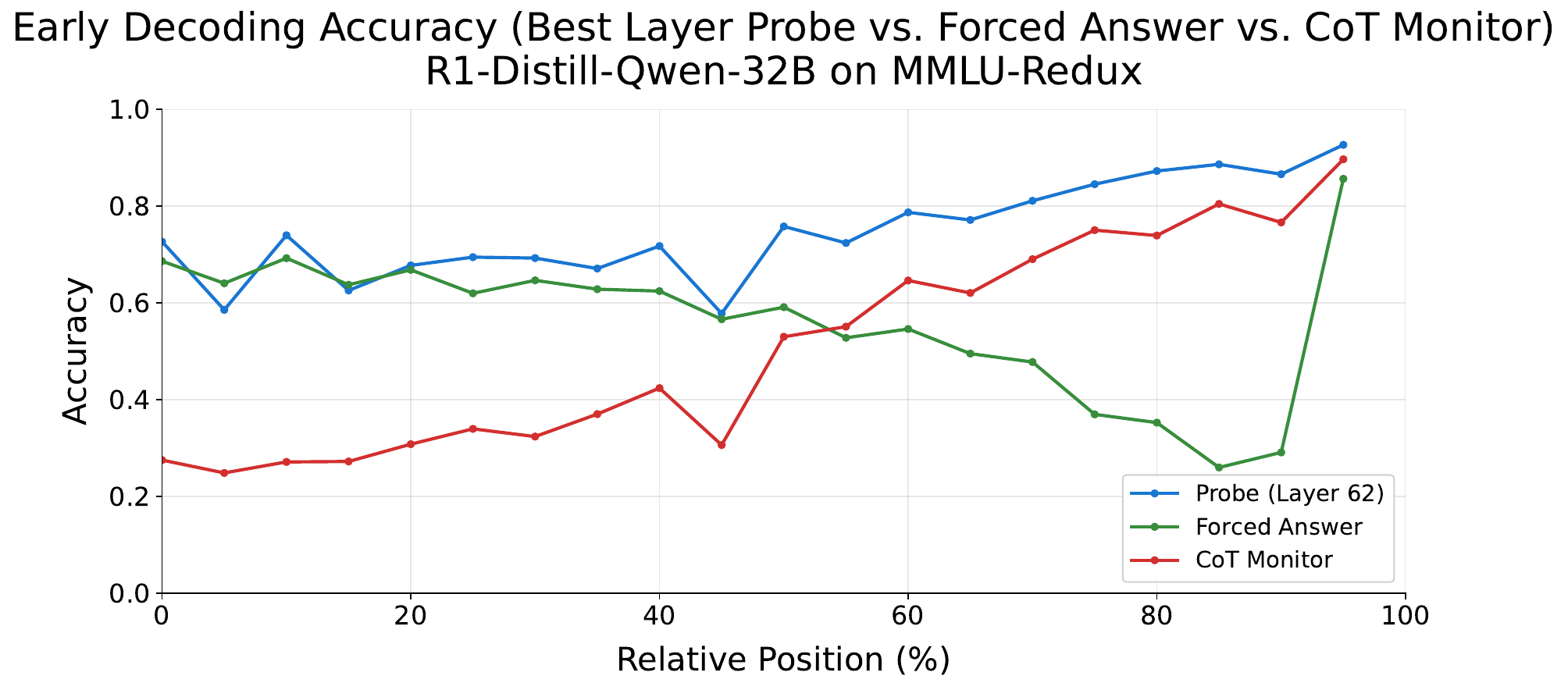}
      \caption{\textbf{Probe, forced answering, and CoT monitor accuracy by reasoning position for DeepSeek-R1-Distill-32B on MMLU.} We see a similar trend to the 14B model, with forced answering accuracy dropping in the second half of reasoning. Both probe accuracy and forced answering accuracy remain the same at the start, implying that in-weights knowledge does not improve significantly between these model sizes.}
      \label{fig:distill-32b-three-methods}
  \end{figure*}

\FloatBarrier
\section{Attention Probe vs. Forced Answer over Sequence} \label{appendix-probe-vs-forced}

\begin{figure*}[h!]
    \centering
    \includegraphics[width=0.9\textwidth]{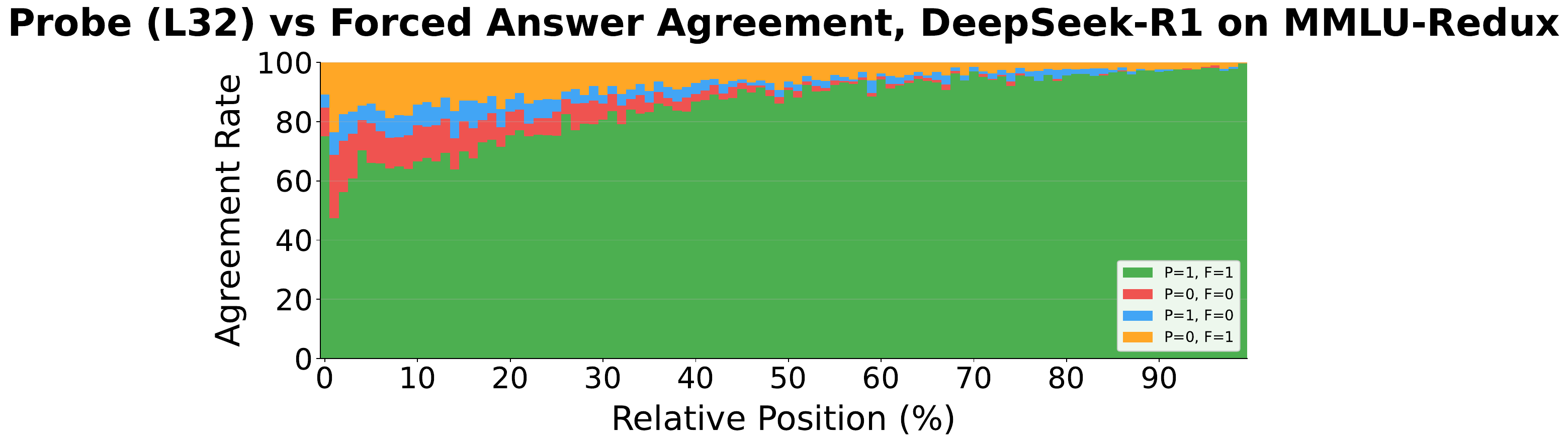}
    \caption{\textbf{Agreement of the best performing probe layer and forced answering for DeepSeek-R1 on MMLU at position bins of the sequence} At the beginning of the sequence, forced answering is correct more often than probes when they disagree. This difference diminishes over reasoning.}
    \label{fig:forced-agreement-mmlu-r1}
\end{figure*}

\begin{figure*}[h!]
    \centering
    \includegraphics[width=0.9\textwidth]{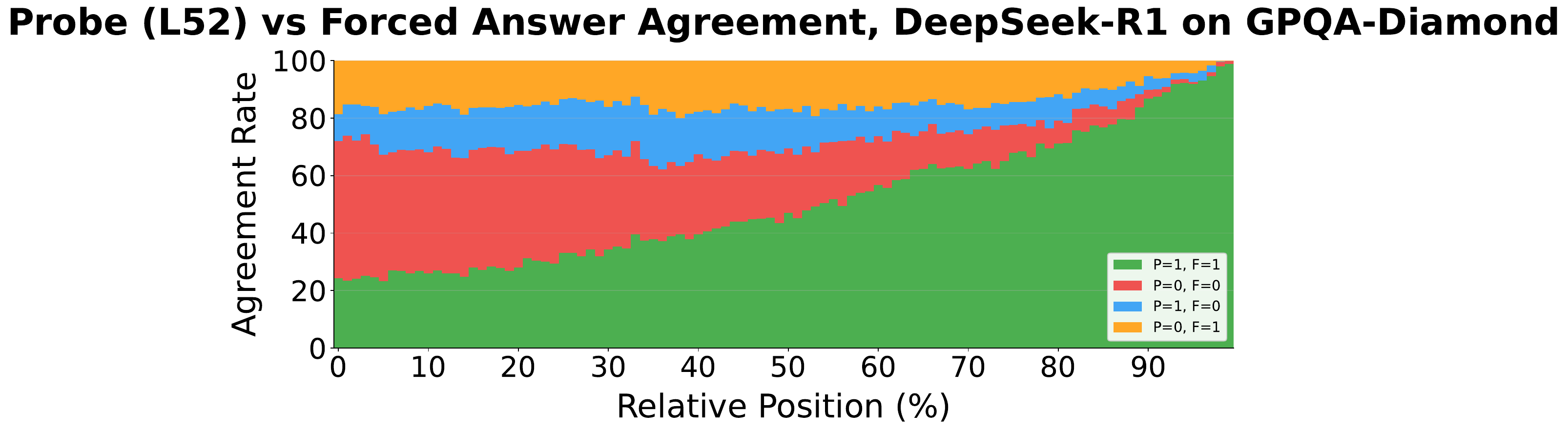}
    \caption{\textbf{Agreement of the best performing probe layer and forced answering for DeepSeek-R1 on GPQA-D at position bins of the sequence.} In GPQA-D, both methods are incorrect 40\% of the time at the beginning of reasoning. When they disagree, forced answering is correct slightly more often.}
    \label{fig:forced-agreement-gpqa-r1}
\end{figure*}

\begin{figure*}[h!]
    \centering
    \includegraphics[width=0.9\textwidth]{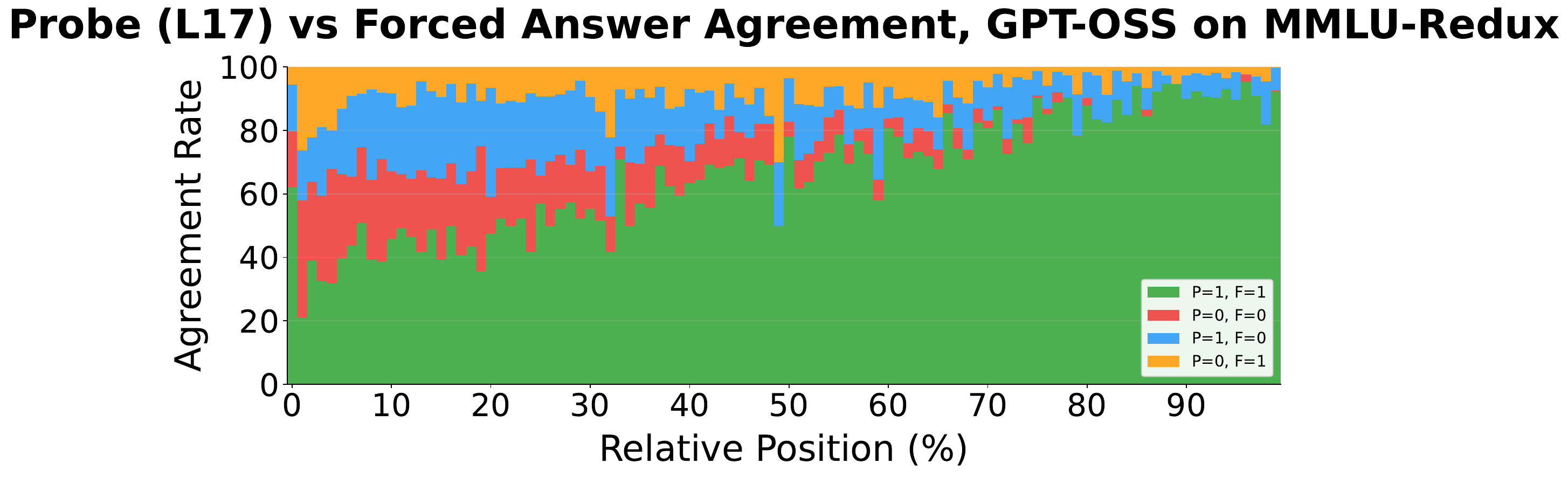}
    \caption{\textbf{Agreement of the best performing probe layer and forced answering for GPT-OSS on MMLU at position bins of the sequence.} We see a less smooth trend for GPT-OSS as responses are shorter. Here, probes outperform forced answering when they disagree.}
    \label{fig:forced-agreement-mmlu-gpt-oss}
\end{figure*}

\begin{figure*}[h!]
    \centering
    \includegraphics[width=0.9\textwidth]{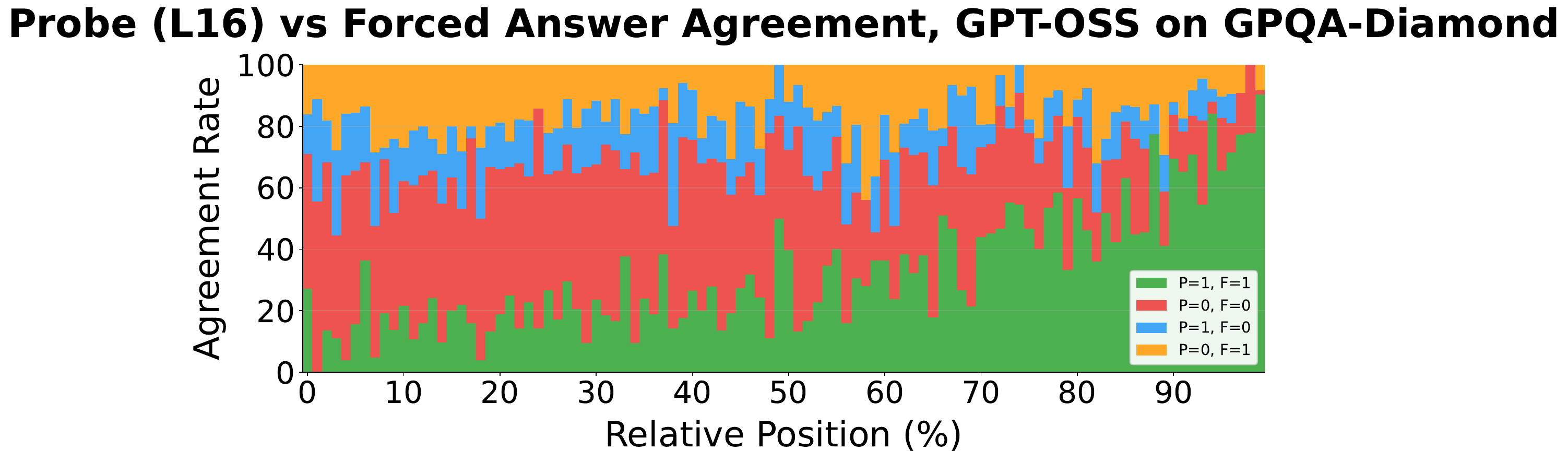}
    \caption{\textbf{Agreement of the best performing probe layer and forced answering for GPT-OSS on GPQA-D at position bins of the sequence.} We see a similar trend to DeepSeek-R1, with both methods being incorrect more often than either is correct until the end of reasoning. Here, forced answering outperforms probing when they disagree, which could be a result of poor probe transfer from MMLU.}
    \label{fig:forced-agreement-gpqa-gpt-oss}
\end{figure*}

We plot the agreement of the best performing probe layer and forced answer at the same step, in 1\% sequence length bins for both of our models and datasets below to show dynamics in Figures~\ref{fig:forced-agreement-mmlu-r1}, \ref{fig:forced-agreement-gpqa-r1}, \ref{fig:forced-agreement-mmlu-gpt-oss} and \ref{fig:forced-agreement-gpqa-gpt-oss}. 

\FloatBarrier

\section{GPT-OSS Calibration for Early Exit} \label{appendix-gpt-oss-calibration-early-exit}

\begin{figure}[!h]
    \centering
    \includegraphics[width=0.5\columnwidth]{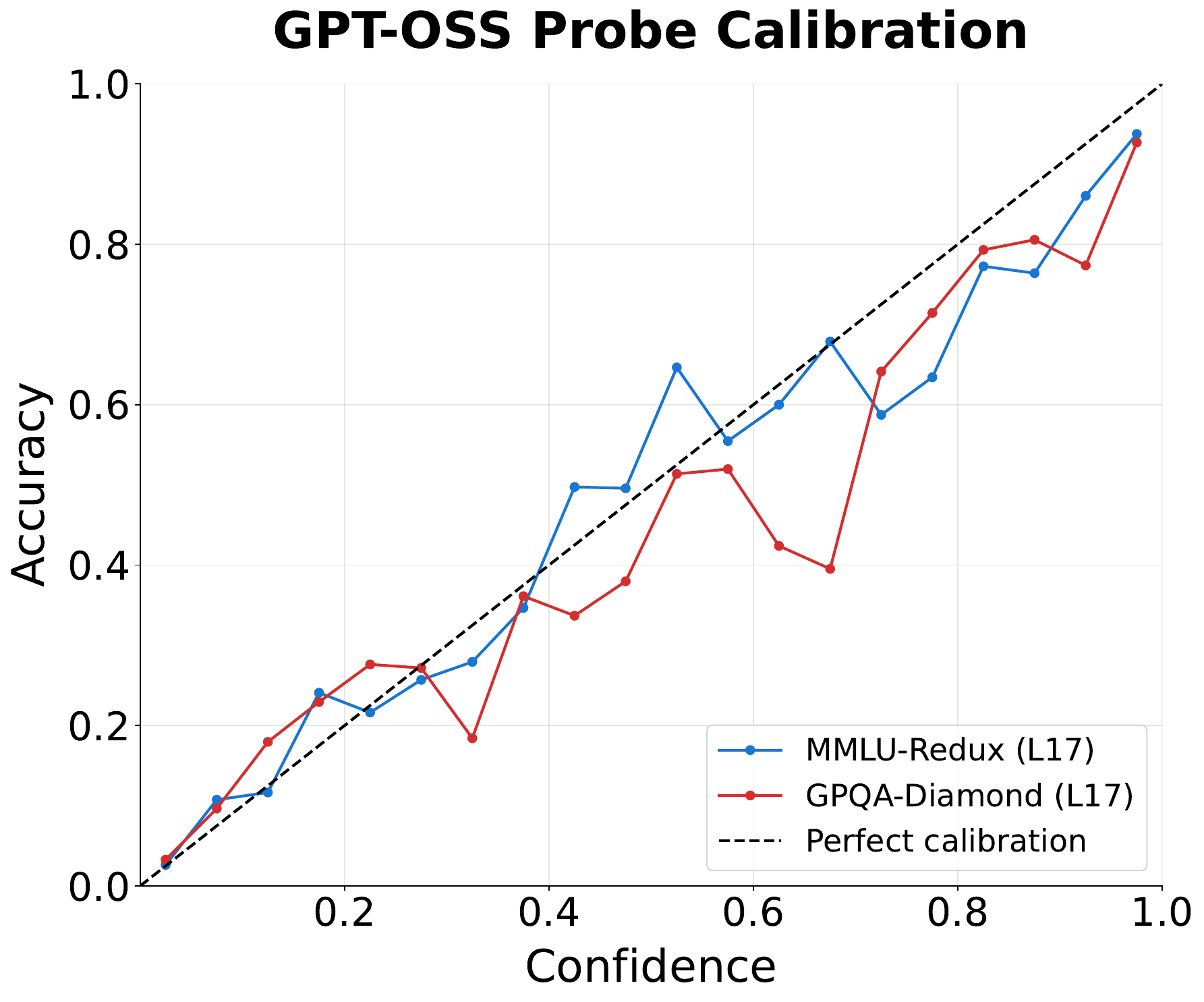}
    \caption{\textbf{Accuracy vs.\ confidence threshold.} GPT-OSS probes for MMLU are well-calibrated, staying close to the perfect calibration line. We observe that when transferring to GPQA-D, probes are more overconfident, potentially due to direct transfer.}
    \label{fig:accuracy-vs-confidence-oss}
\end{figure}

\begin{figure}[!h]
    \centering
    \includegraphics[width=0.6\columnwidth]{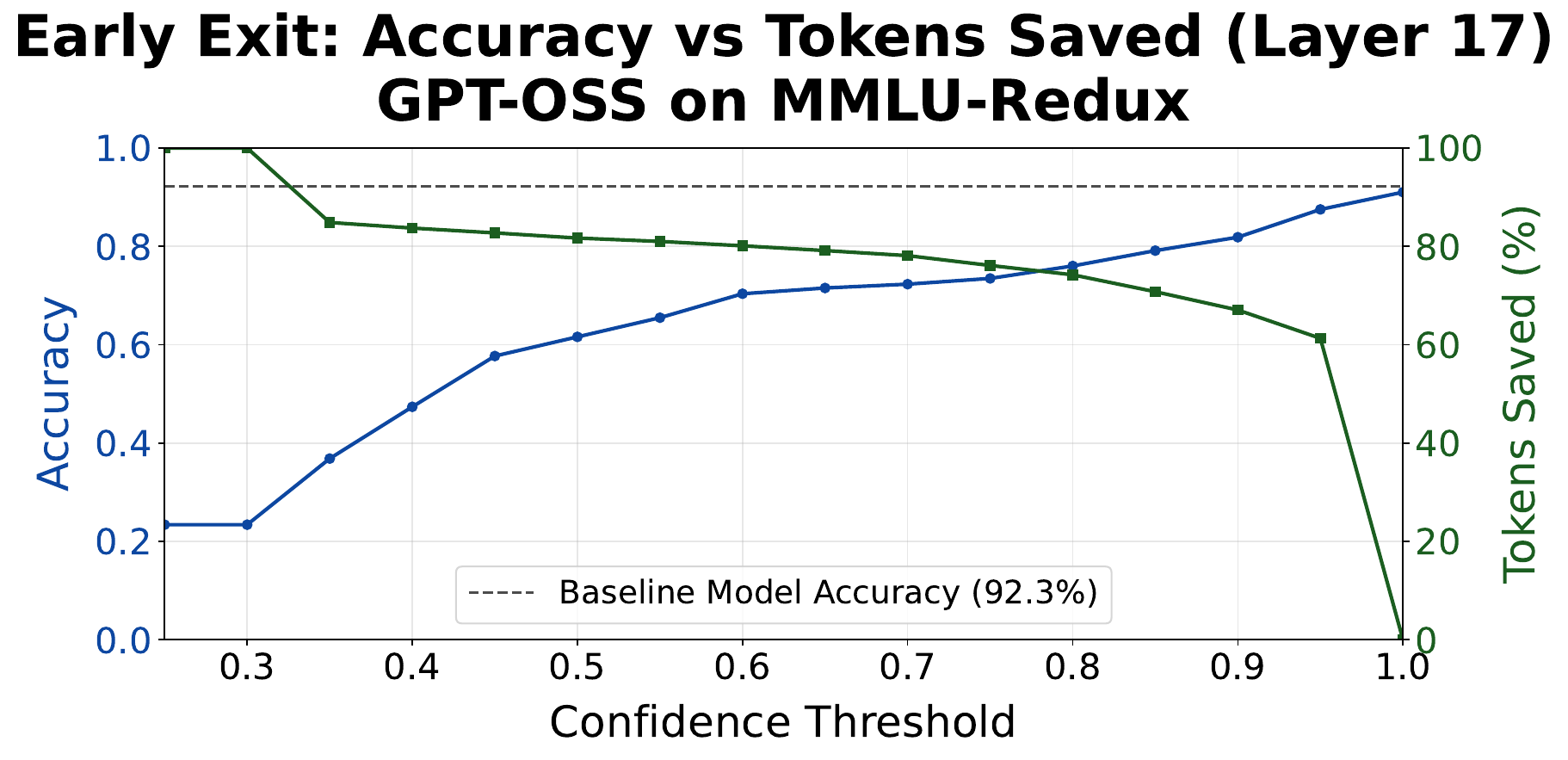}
    \caption{\textbf{Exit position vs.\ confidence threshold.} We observe similar trends for GPT-OSS and DeepSeek-R1 on MMLU, where the majority of tokens can be saved by early exiting with 90\% confidence while having a marginal drop in performance.}
    \label{fig:exit-vs-confidence1oss}
\end{figure}

\begin{figure}[!h]
    \centering
    \includegraphics[width=0.6\columnwidth]{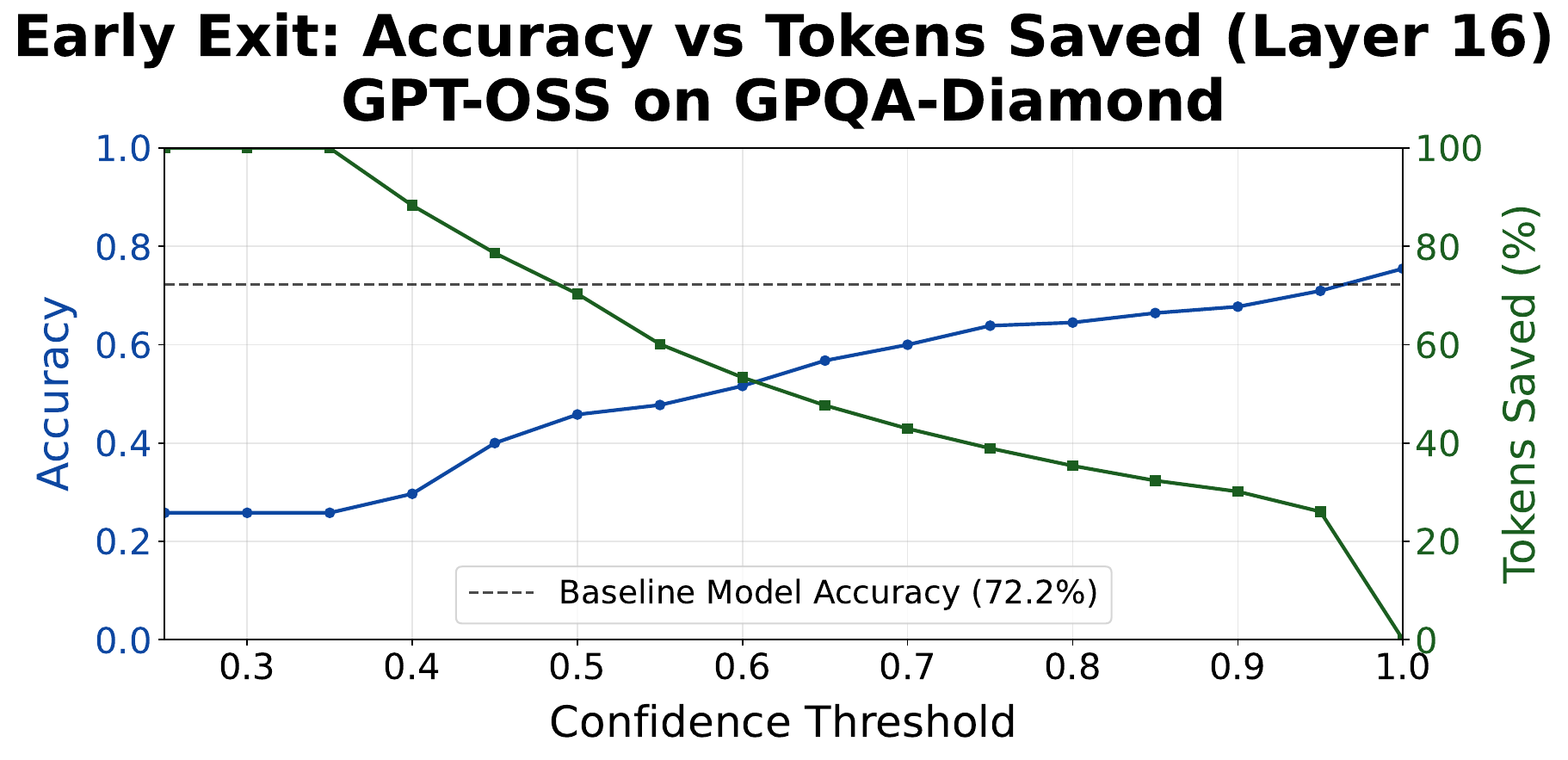}
    \caption{\textbf{Exit position vs.\ confidence threshold.} We again see a similar trend to DeepSeek-R1, with less token saving per accuracy drop. Interestingly, our probe outperforms the original model's accuracy at the end of reasoning.}
    \label{fig:exit-vs-confidence2oss}
\end{figure}

We include the plots for GPT-OSS probe calibration for early exit in Figures \ref{fig:accuracy-vs-confidence-oss}, \ref{fig:exit-vs-confidence1oss} and \ref{fig:exit-vs-confidence2oss}, mirroring the DeepSeek-R1 calibration results in Section~\ref{sec:early-exit}. 

\FloatBarrier

\section{Additional Inflection Analysis} \label{appendix-additional-probe-inflection}

We include analogous inflection point analysis to Section~\ref{sec:results_inflections} for GPT-OSS responses on MMLU. Table~\ref{tab:reasoning-events-confidence-gptoss} shows that the same pattern observed for DeepSeek-R1 holds: high-confidence traces (124/514) produce fewer inflection points per step than other traces (0.053 vs.\ 0.096), consistent with inflections reflecting genuine uncertainty rather than performative reasoning. Notably, GPT-OSS exhibits a higher overall inflection rate than DeepSeek-R1 (0.089 vs.\ 0.038 per step), driven primarily by a much higher rate of reconsiderations (0.064 vs.\ 0.028 per step).

  \begin{table}[h!]
    \caption{\textbf{Inflection points by probe confidence level for GPT-OSS on MMLU (probe layer 35).} The same pattern holds: high-confidence traces contain fewer inflections per step.}
    \label{tab:reasoning-events-confidence-gptoss}
    \begin{center}
      \begin{footnotesize}
        \begin{sc}
          \begin{tabular}{lccc}
            \toprule
            Event Type
            & \begin{tabular}[c]{@{}c@{}} High \\ Conf. \end{tabular}
            & \begin{tabular}[c]{@{}c@{}} Non-High \\ Conf. \end{tabular} \\
            \midrule
            Reconsideration & 0.043 & 0.068 \\
            Realization      & 0.002  & 0.012  \\
            Backtrack     & 0.007 & 0.016  \\
            \midrule
            \textbf{Total} & \textbf{0.053} & \textbf{0.096} \\
            \bottomrule
          \end{tabular}
        \end{sc}
    \end{footnotesize}
    \end{center}
    \vskip -0.1in
  \end{table}

\begin{figure}[ht]
    \centering
    \includegraphics[width=0.6\columnwidth]{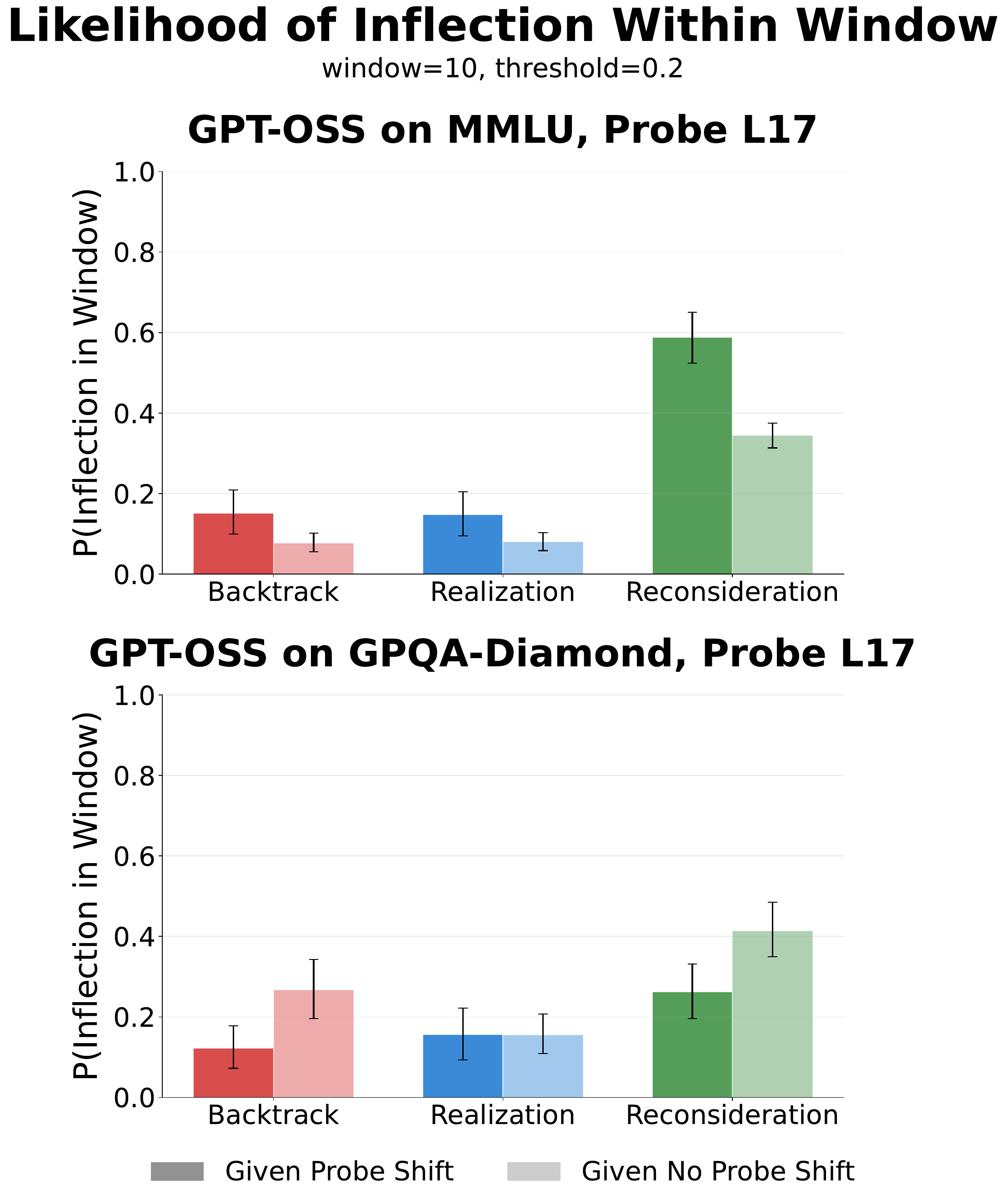}
    \caption{\textbf{Rate of inflection points occurring within windows beginning with probe shifts and windows containing no probe shifts.} We observe that reconsideration points occur twice as often for 20\% confidence shifts of the highest probability answer choice for MMLU with a window size of 10 steps ahead, but find no other significant trend for other answer types or for GPQA-D.}
    \label{fig:gpt_oss_inflection_probe_correlation}
\end{figure}

Figure~\ref{fig:gpt_oss_inflection_probe_correlation} shows the likelihood of each inflection type occurring within a 10-step window following a 20\% probe confidence shift. On MMLU, reconsiderations are nearly
  twice as likely to follow a probe shift (59\%) than to occur without one (35\%), and backtracks and realizations show a similar pattern. This suggests that probe confidence shifts predict upcoming inflection points. However, on GPQA-Diamond, we see the opposite trend, with greater likelihood of inflection given no probe shift in the window.

\subsection{Probe Shifts and Inflection Point Correlation}                                                                                                                                                         
  \label{sec:probe-shift-inflection-correlation}                                                                                                                                                                                                               
  We investigate the temporal relationship between probe confidence shifts and inflection points in the reasoning trace. For each confidence shift threshold (the minimum change in the probe's top predicted
  probability) and window size (the number of subsequent steps to search), we compute two conditional probabilities: (1) $P(\text{inflection follows} \mid \text{probe shift}) - P(\text{inflection follows} \mid    
  \text{no probe shift})$, measuring whether probe shifts predict upcoming inflections, and (2) the reverse, measuring whether inflections predict subsequent probe shifts.

  \begin{figure*}[ht]
      \centering
      \includegraphics[width=\textwidth]{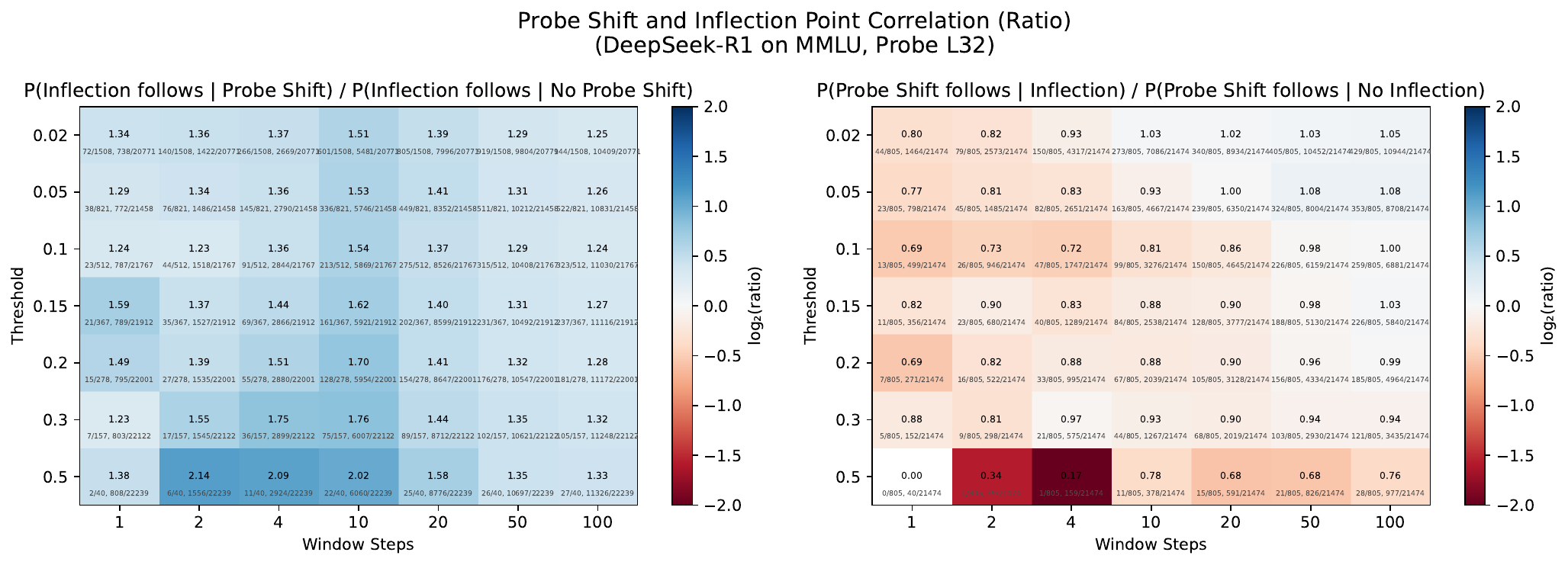}
      \caption{\textbf{Probe shift and inflection point correlation for DeepSeek-R1 on MMLU (probe layer 60).} Left: inflection points are substantially more likely to follow probe confidence shifts, with the effect strongest at a high threshold (0.5) and moderate window sizes (2--10 steps), being twice as likely. Right: inflection points show either no correlation or a negative correlation depending on the window and threshold.}
      \label{fig:r1-probe-inflection-heatmap}
  \end{figure*}

  \begin{figure*}[ht]
      \centering
      \includegraphics[width=\textwidth]{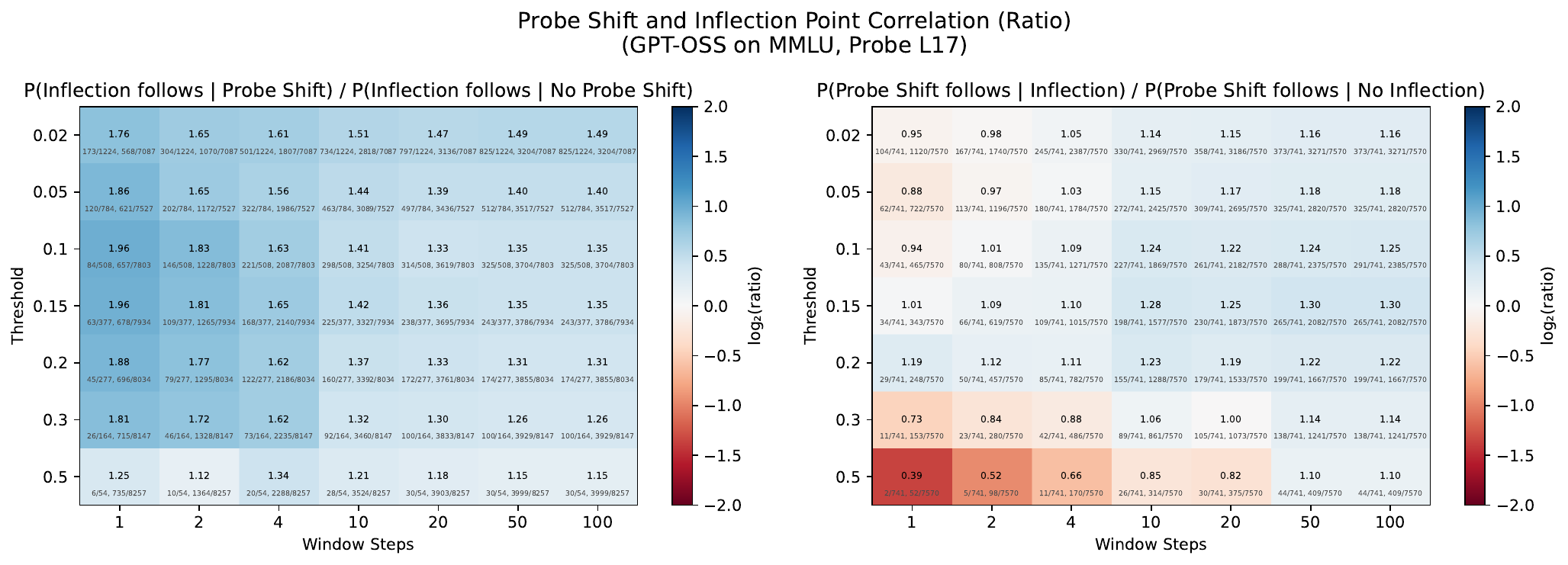}
      \caption{\textbf{Probe shift and inflection point correlation for GPT-OSS on MMLU (probe layer 35).} We observe a different pattern of order here: probe shifts predict inflections (left), mainly with a window size of 1 while inflections also predict probe shifts for some window/threshold combinations. Howver, this trend is inconsistent, complicating analysis particularly in smaller window sizes.}
      \label{fig:gptoss-probe-inflection-heatmap}
  \end{figure*}

\clearpage


\end{document}